\pdfoutput=1

\documentclass[11pt]{article}
\usepackage[dvipsnames]{xcolor}

\usepackage[preprint]{acl}

\usepackage{times}
\usepackage{latexsym}

\usepackage[T1]{fontenc}

\usepackage[utf8]{inputenc}

\usepackage{microtype}

\usepackage{inconsolata}

\usepackage{graphicx}
\usepackage{tabularx}
\usepackage{multirow}
\usepackage{amsfonts}
\usepackage{amssymb}
\usepackage{xspace}
\usepackage{lipsum}
\usepackage{adjustbox}
\usepackage{booktabs}
\usepackage{makecell}
\usepackage{url}
\usepackage{enumitem}
\setlist[enumerate]{topsep=0pt, partopsep=0pt, parsep=0pt, itemsep=0pt}
\setlist[itemize]{topsep=0pt, partopsep=0pt, parsep=0pt, itemsep=0pt}
\usepackage{listings}
\usepackage{xcolor}
\usepackage{inconsolata}
\usepackage[most]{tcolorbox}
\usepackage{caption}
\usepackage{float}
\usepackage{pgfplots}
\pgfplotsset{compat=1.17}
\usepackage{pgfplotstable}
\usepackage{amsthm}
\usetikzlibrary{calc}

\usepackage{xcolor}
\usepackage{tikz}
\usetikzlibrary{arrows.meta, positioning, shapes.geometric, calc}
\usepackage{pgfplots}
\usepgfplotslibrary{groupplots}
\pgfplotsset{compat=1.18}

\definecolor{aggcolor}{HTML}{133167}
\definecolor{conscolor}{HTML}{b2cfe3}
\definecolor{graynode}{HTML}{d9d9d9}
\definecolor{grayline}{HTML}{999999}
\definecolor{failred}{HTML}{c0392b}
\definecolor{blueLight}{HTML}{B6CDE0}
\definecolor{blueDark}{HTML}{1F3563}
\definecolor{grayBorder}{HTML}{888780}
\definecolor{fig8highcolor}{HTML}{133167}
\definecolor{fig8lowcolor}{HTML}{b2cfe3}

\definecolor{dsagg}{HTML}{1F3A6E}
\definecolor{dscons}{HTML}{A9C6E8}
\definecolor{gemagg}{HTML}{B5651D}
\definecolor{gemcons}{HTML}{F4C28F}
\definecolor{zhanweicolor}{RGB}{200,40,40}

\pgfplotsset{
  fig34style/.style={
    ybar, bar width=6pt,
    width=8.9cm, height=7cm,
    symbolic x coords={fo1,fom,fo2,gap1,ns1,nsm,ns2,gap2,ws1,wsm,ws2},
    xtick={fo1,fo2,ns1,ns2,ws1,ws2},
    xticklabels={,,,,,},
    title style={font=\normalsize},
    y tick label style={font=\small},
    nodes near coords,
    nodes near coords style={font=\small, rotate=90, anchor=west, inner sep=2pt},
    enlarge x limits=0.06,
    tick align=outside,
    axis line style={-},
    ymajorgrids, grid style={gray!15},
    clip=false,
  }
}

\pgfplotsset{
  fig8style/.style={
    ybar, bar width=13pt,
    width=10.5cm, height=7cm,
    symbolic x coords={c1,c2,c3,c4},
    xtick=data,
    xticklabels={DeepSeek -- no stats, DeepSeek -- w/ stats, Gemini -- no stats, Gemini -- w/ stats},
    x tick label style={font=\small, rotate=20, anchor=east},
    y tick label style={font=\small},
    nodes near coords, nodes near coords style={font=\small},
    enlarge x limits=0.2,
    ymin=0, ymax=115,
    ylabel={Assumption Rate before Planning (\%)},
    tick align=outside,
    ymajorgrids, grid style={gray!15},
  }
}

\definecolor{fig3cAggPlanner}{HTML}{0a1f44}    
\definecolor{fig3cConsPlanner}{HTML}{2e5c94}   
\definecolor{fig3cAggFormalizer}{HTML}{7ba3cc} 
\definecolor{fig3cConsFormalizer}{HTML}{c5dbef} 
 
\pgfplotsset{
  fig3cstyle/.style={
    width=4.3cm, height=4.6cm,
    xlabel={Attempt \#},
    xlabel style={font=\scriptsize},
    ylabel style={font=\small},
    xmin=0.5, xmax=12.5,
    ymin=0, ymax=105,
    xtick={2,6,10},
    tick align=outside,
    y tick label style={font=\tiny},
    x tick label style={font=\tiny},
    title style={font=\scriptsize},
    ymajorgrids, grid style={gray!15},
    mark options={scale=0.8},
  }
}

\pgfplotsset{
  aggplanner/.style={color=fig3cAggPlanner, mark=o, thick, mark size=1.7pt},
  consplanner/.style={color=fig3cConsPlanner, mark=square, thick, mark size=1.7pt},
  aggformalizer/.style={color=fig3cAggFormalizer, mark=triangle, thick, mark size=2pt},
  consformalizer/.style={color=fig3cConsFormalizer, mark=diamond, thick, mark size=2pt},
}

\newcommand{\drexel}{%
  \hspace{1pt}
  \begingroup\normalfont
  \includegraphics[height=1.3\fontcharht\font`\B]{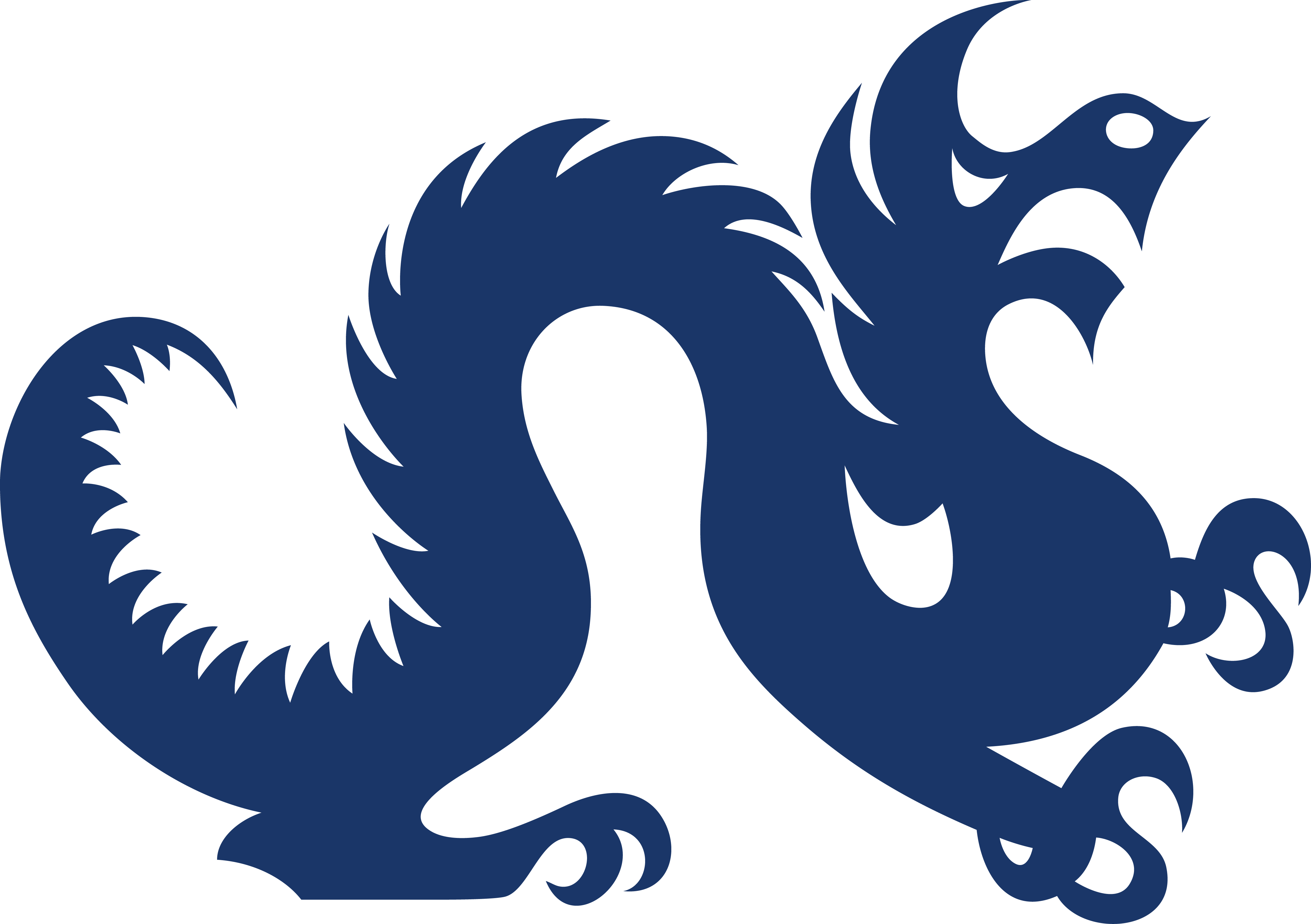}%
  \endgroup
  \hspace{1pt}
}
\newcommand{\northeastern}{%
  \hspace{1pt}
  \begingroup\normalfont
  \includegraphics[height=1.3\fontcharht\font`\B]{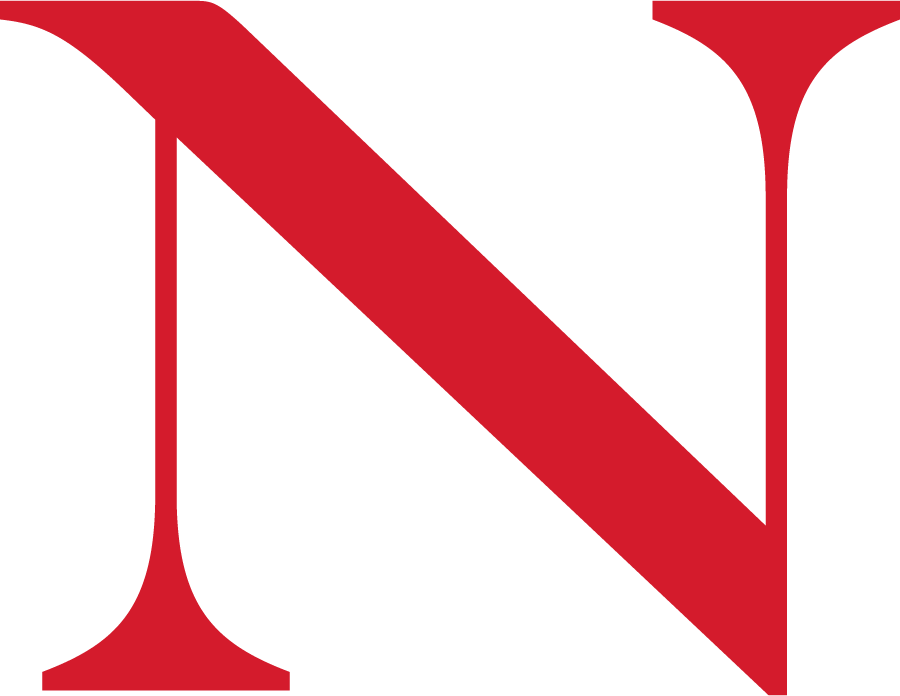}%
  \endgroup
  \hspace{1pt}
}

\newcommand{\uci}{%
  \hspace{1pt}
  \begingroup\normalfont
  \includegraphics[height=1.3\fontcharht\font`\B]{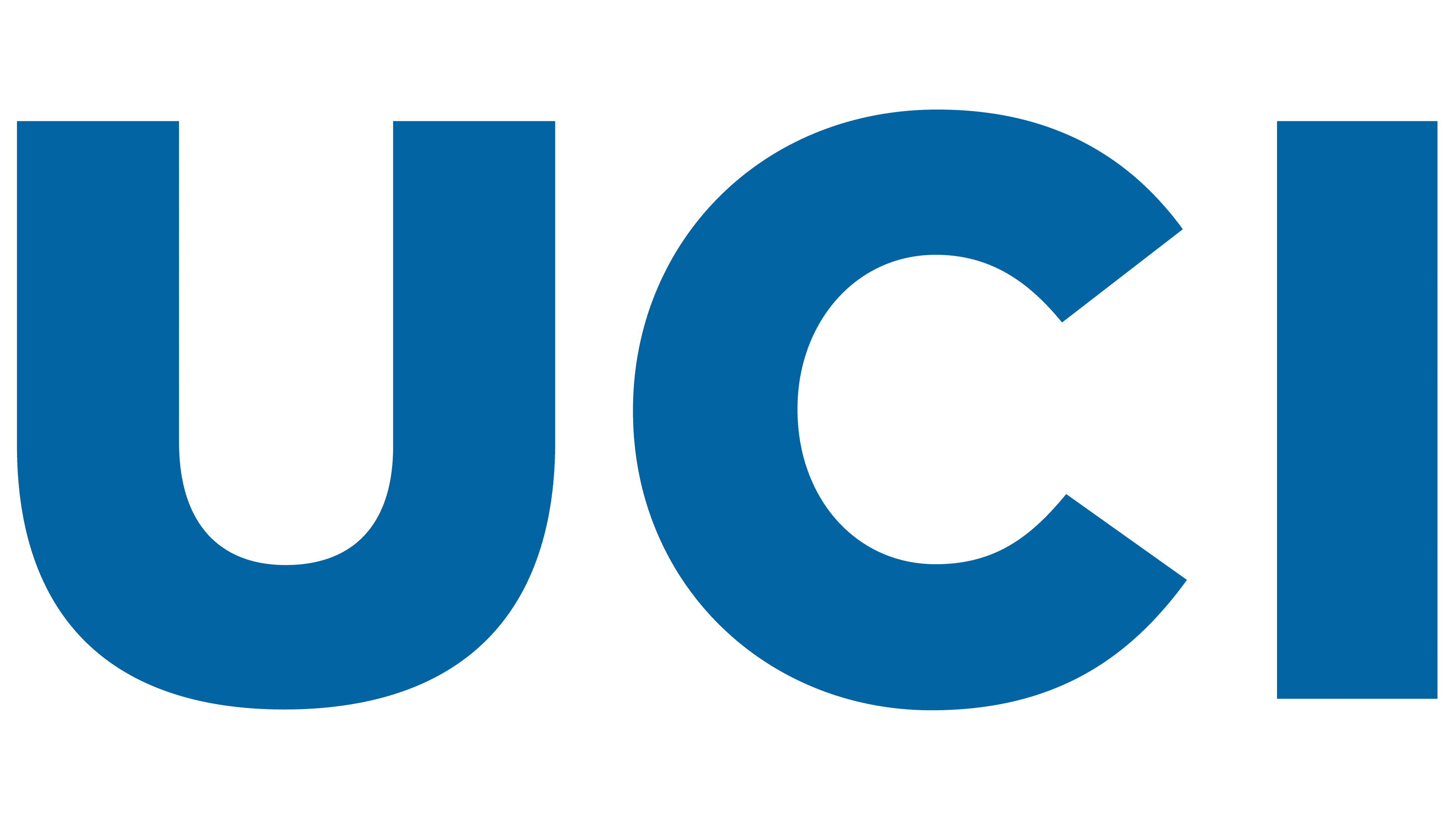}%
  \endgroup
  \hspace{1pt}}

\newcommand{\penn}{%
  \hspace{1pt}
  \begingroup\normalfont
  \includegraphics[height=1.3\fontcharht\font`\B]{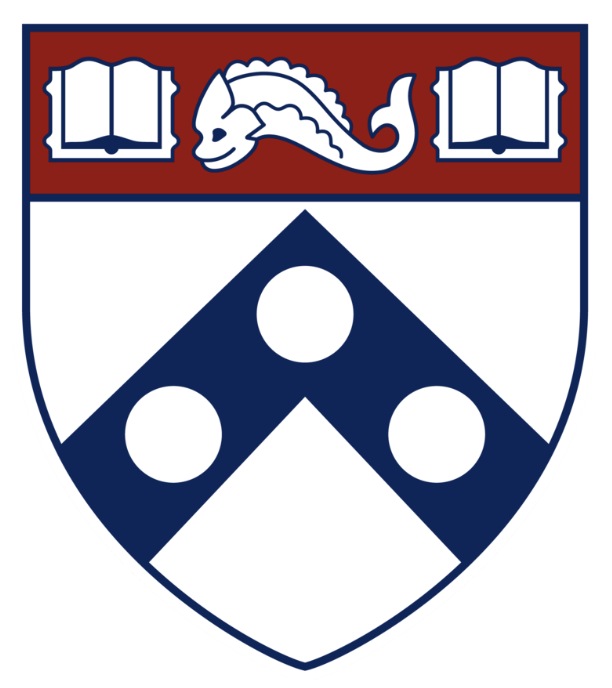}%
  \endgroup
  \hspace{1pt}}

\newcolumntype{Y}{>{\centering\arraybackslash}X}

\lstdefinestyle{mystyle}{
    basicstyle=\ttfamily\small,
    breaklines=true,
    frame=single,
    backgroundcolor=\color{gray!5},
    keywordstyle=\color{black},
    commentstyle=\color{gray},
    stringstyle=\color{black},
    showstringspaces=false,
    tabsize=2,
    rulecolor=\color{black},
    framesep=6pt, 
    columns=fullflexible,
}
\title{Planning Commitment of LLM under Partial Observability}
\author{
  {\bfseries Cassie Huang\drexel \enspace Zhanwei Cao\northeastern \enspace Yeawon Lee\drexel} \enspace Nathalia Gomez\drexel \enspace Zhuoyi Yang\uci\\[0.2em]
  {\bfseries Ian G. Harris\uci \enspace Tiffany D. Do\drexel \enspace Jiayuan Mao\penn \enspace Li Zhang\drexel} \\[0.3em]
  \drexel Drexel University \hspace{8pt} \northeastern Northeastern University \\
  \hspace{8pt} \uci University of California, Irvine \penn University of Pennsylvania\\[0.2em]
  {\tt \{Cassie.Huang|Harry.Zhang\}@drexel.edu}
}

\begin{document}
\maketitle
\begin{abstract}

LLM-based planning agents commit to plans of different lengths before interacting with a partially observable environment. Some aggressively commit to a long plan based on assumptions about unobserved states, while others conservatively plan only over what they have observed. Both behaviors emerge implicitly in pre-trained models, whether the LLM outputs a plan directly or generates a PDDL formulation for a classical solver. Aggressive commitment risks costly failures, while conservative commitment risks excessive LLM calls. This choice pits safety against efficiency, a trade-off that has been neither characterized systematically nor made controllable. We investigate aggressive and conservative LLM planners and formalizers on two partially observable domains, where priors are provided as either explicit statistics or common-sense knowledge. Across three LLMs, both strategies reach similar accuracy. Conservative methods achieve fewer failed attempts (safety) at the cost of more API calls (efficiency). Aggressive methods fail more often but replan efficiently only when priors are informative. Interestingly, analysis shows that LLMs treat the provided priors nearly as facts, which risks failure when priors are unreliable. We also explore hybrid strategies with tunable hyperparameters that adjust the balance between safety and efficiency.


\end{abstract}

\section{Introduction}

Large language model agents have been used for planning in two ways. In the LLM-as-planner paradigm, the model directly generates action plans from natural-language goals and observations, offering flexibility in open-ended environments and the ability to adapt rapidly when execution deviates from expectation \cite{wei-etal-2025-plangenllms}. In the LLM-as-formalizer paradigm, the model instead translates natural-language tasks and observations into an explicit symbolic representation that is solved by an external planner, offering much more formal guarantee, trust, and consistency especially desired in high-stakes applications \cite{tantakoun-etal-2025-llms}. Under partial observability, these paradigms are typically extended through iterative interaction: planner-based agents interleave reasoning, acting, and replanning as new observations arrive, while formalizer-based agents incrementally update their symbolic model and invoke the solver. 

How long a plan should an LLM commit to before each execution? In the LLM-as-planner paradigm, this choice is often implicit: some approaches encourage aggressive commitment by generating an overall plan upfront and relying on replanning when execution encounters unexpected obstacles \cite{erdogan2025planandact}, whereas others adopt a more conservative strategy by interleaving reasoning, acting, and observation at each step \cite{yao2023react}. In contrast, the LLM-as-formalizer paradigm is often bound by the closed-world assumption of formal frameworks like PDDL/STRIPS \cite{FikesNilsson1971}, making it difficult to represent uncertain beliefs directly, so existing iterative formalization approaches primarily operate conservatively by formalizing and planning over observed information \cite{gong-etal-2026-iterative}. Each choice has a cost: aggressive commitment risks failed actions when assumptions prove wrong, compromising \textit{safety}, while conservative commitment requires more rounds of LLM calls, compromising \textit{efficiency}. Despite the prevalence of these behaviors across LLM planning systems, planning commitment and its safety–efficiency trade-off have been neither systematically studied nor made controllable.


\begin{figure*}[t!]
    \centering
    \includegraphics[width=\linewidth]{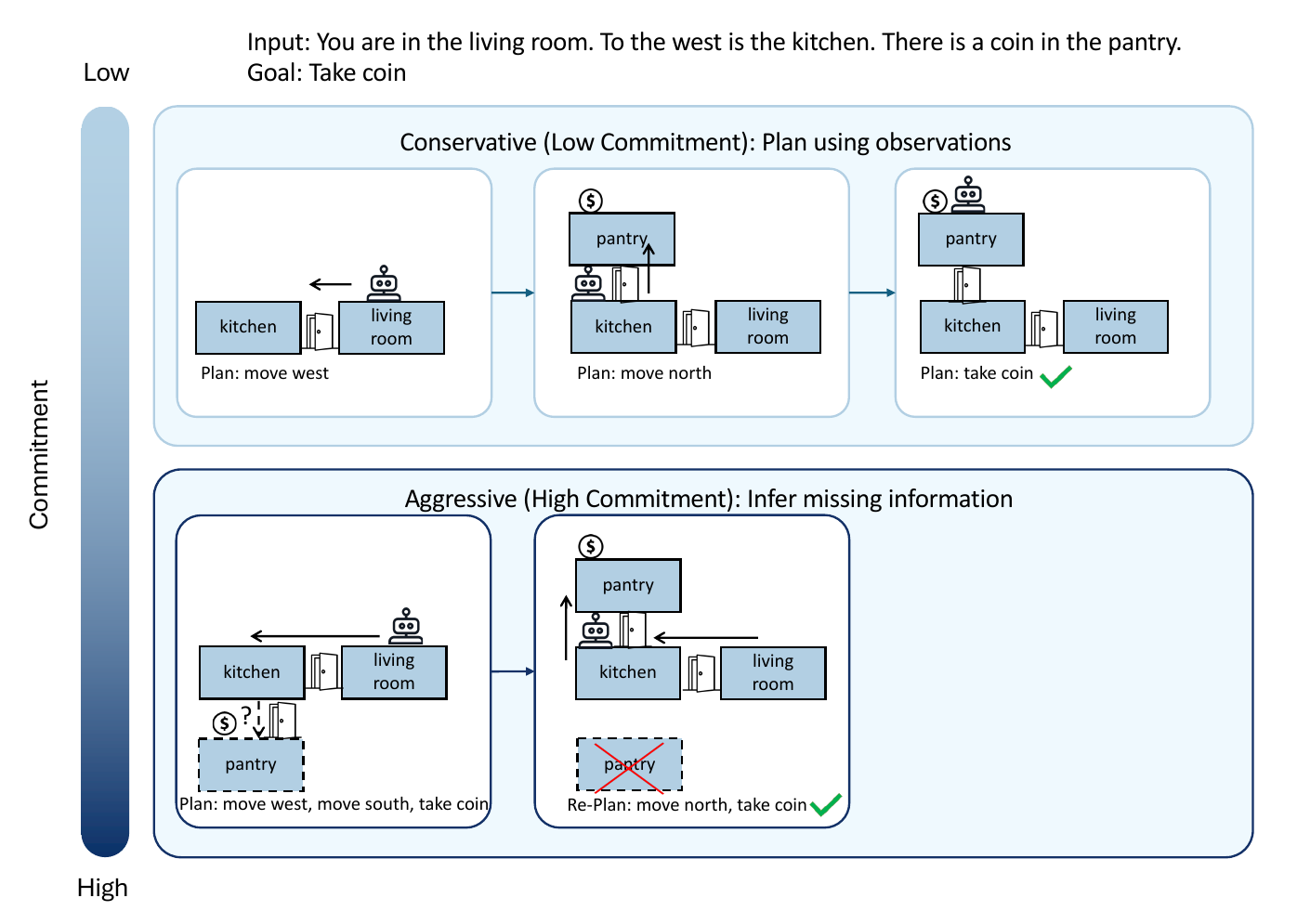}
    \caption{In partially observable environments, models are required to either only make plans from observations when interacting with the environment (Conservative) or assume missing information and replan based on environment interactions (Aggressive).}
    \label{fig:commitment}
\end{figure*}

We take the first step toward studying the planning commitment of LLMs by constructing two controlled, partially observable, text-based environments built on well-known domains of navigation and object manipulation. For Coin Collector \cite{yuan2018counting}, we construct the floor plans to follow certain patterns and supply curated statistical priors (e.g., \emph{83\% of the time, room\_3 is to the west of room\_8}) in a domain with arbitrary layouts. For ALFWorld \cite{shridhar2021alfworld}, we design realistic room contents that let the model draw on its own commonsense knowledge. On both domains, we perform a four-way evaluation of aggressive and conservative LLM planners and formalizers. We instantiate each combination with a representative design from prior work: the conservative planner interleaves reasoning and acting as in ReAct \cite{yao2023react}, the aggressive planner generates a full plan and repairs it as in Plan-and-Act \cite{erdogan2025planandact}, and the conservative formalizer iteratively grows its PDDL from observations as in PDDLego+ \cite{gong-etal-2026-iterative}. We introduce an aggressive formalizer that instead writes assumed facts directly into the PDDL problem, letting us test whether formalizers can benefit from committing beyond what they observe. We find that when priors or common sense are available, aggressive methods are more efficient, while conservative methods are usually safer, with far fewer failed attempts. We also find that aggressive formalizers treat provided priors nearly as fact, and that hybrid strategies offer a middle ground between the two. From these findings, we recommend a level of commitment depending on priorities and environment, ranging from low-stakes, well-characterized settings to high-stakes, unfamiliar ones.

We find that both strategies reach similar success rates but differ sharply in cost: conservative methods are safer, with far fewer failed attempts, while aggressive methods are more efficient when priors or common sense are available. We also find that LLMs treat provided priors nearly as facts, which risks failure when priors are unreliable. Finally, hybrid strategies with tunable hyperparameters let practitioners control the balance between safety and efficiency. From these findings, we recommend aggressive commitment for low-stakes, well-characterized settings and conservative commitment for high-stakes, unfamiliar ones.

Our contributions are as follows:
\begin{itemize}
    \item We frame \textbf{planning commitment} as a design axis independent of the planner--formalizer distinction, and introduce an aggressive formalizer to complete a four-way comparison.
    \item We build two partially observable environments, with priors given as explicit statistics, as common sense, or not at all.
    \item We show that aggressive and conservative commitment reach similar success rates but trade safety for efficiency, and that LLMs treat probabilistic priors nearly as facts.
    \item We show that hybrid strategies with tunable hyperparameters control this trade-off, and derive practical recommendations for choosing a level of commitment.
\end{itemize}

\section{Related Work}
\label{sec:related}

\paragraph{Conservative LLM planners.}
ReAct interleaves reasoning with environment actions and incorporates each observation before choosing the next action \cite{yao2023react}. Robotic and
agentic variants such as SayCan, Inner Monologue, Reflexion, and LLaMAR keep
this closed loop \cite{ahn2022saycan,huang2022inner,shinn2023reflexion,nayak2025llamarlonghorizonplanningmultiagent}.
These agents adapt well and gather information as they go, but they incur
repeated inference cost and can behave myopically over long horizons
\cite{wang2026reasoningfailsplanplanningcentric}.

\noindent\textbf{Aggressive LLM planners.}
ProgPrompt, LLM-Planner, AdaPlanner, DEPS, and Plan-and-Act first generate a
long-horizon plan and repair it when execution deviates
\cite{singh2023progprompt,song2023llmplanner,NEURIPS2023_b5c8c1c1,NEURIPS2023_6b8dfb8c,erdogan2025planandact}.
This gives global coherence, but early modeling errors can corrupt entire
plans, and most methods rely on implicit parametric knowledge rather than a
calibrated distribution over unobserved facts.
 
\noindent\textbf{Bounded-lookahead planners.}
LATS, LLM-guided Monte Carlo planning, and FLARE search over short horizons of
hypothesized trajectories before committing to the next action
\cite{zhou2024languageagenttreesearch,shi2025montecarloplanninglarge,hao-etal-2023-reasoning,zhao2023largelanguagemodelscommonsense,wang2026reasoningfailsplanplanningcentric}.
They sit between our two extremes; we treat this intermediate level of
commitment as outside our scope.
 
\noindent\textbf{Conservative LLM formalizers.}
LLM+P and NL2Plan translate fully observable tasks into PDDL for an external
planner \cite{liu2023llm+,gestrin-et-al-icaps2024wshaxp, tantakoun-etal-2025-llms, huang-zhang-2025-limit}. Under
partial observability, PDDLego and PDDLego+ incrementally update the PDDL
from observations \cite{zhang-etal-2024-pddlego,gong-etal-2026-iterative}, and WorldCoder
learns transition models through interaction \cite{tang2024worldcoder}.
These approaches inherit the closed-world assumption of symbolic models,
which makes uncertainty hard to represent.
 
\noindent\textbf{Aggressive LLM formalizers.}
Conformant, contingent, and FOND planners, PPDDL, and POMDP solvers such as
SARSOP, POMCP, and DESPOT support committing under uncertainty, but they
require complete, hand-specified models
\cite{HOFFMANN2006507,3037062.3037072,1661445.1661706,3038546.3038567,mattmueller-etal-icaps2010,Younes2004PPDDL1, inproceedings,NIPS2010_edfbe1af,NIPS2013_c2aee861}.
Few systems pair this machinery with LLM-generated models. Tru-POMDP, closest
to our aggressive formulation, generates explicit hypotheses about unobserved
objects and goals before belief-space search
\cite{tang2026trupomdptaskplanninguncertainty}; LLM-DP maintains a
distribution over sampled worlds instead of a single hypothesis
\cite{dagan2023dynamicplanningllm}; and PO-PDDL learns symbolic POMDP models
from demonstrations \cite{tang2026popddllearningsymbolicpomdps}. None of
these study how the presence or strength of priors affects planning.

\section{Task Formulation and Methodologies}
\label{sec:formulation_and_methodologies}
\begin{figure*}
    \centering
    \includegraphics[width=.8\linewidth]{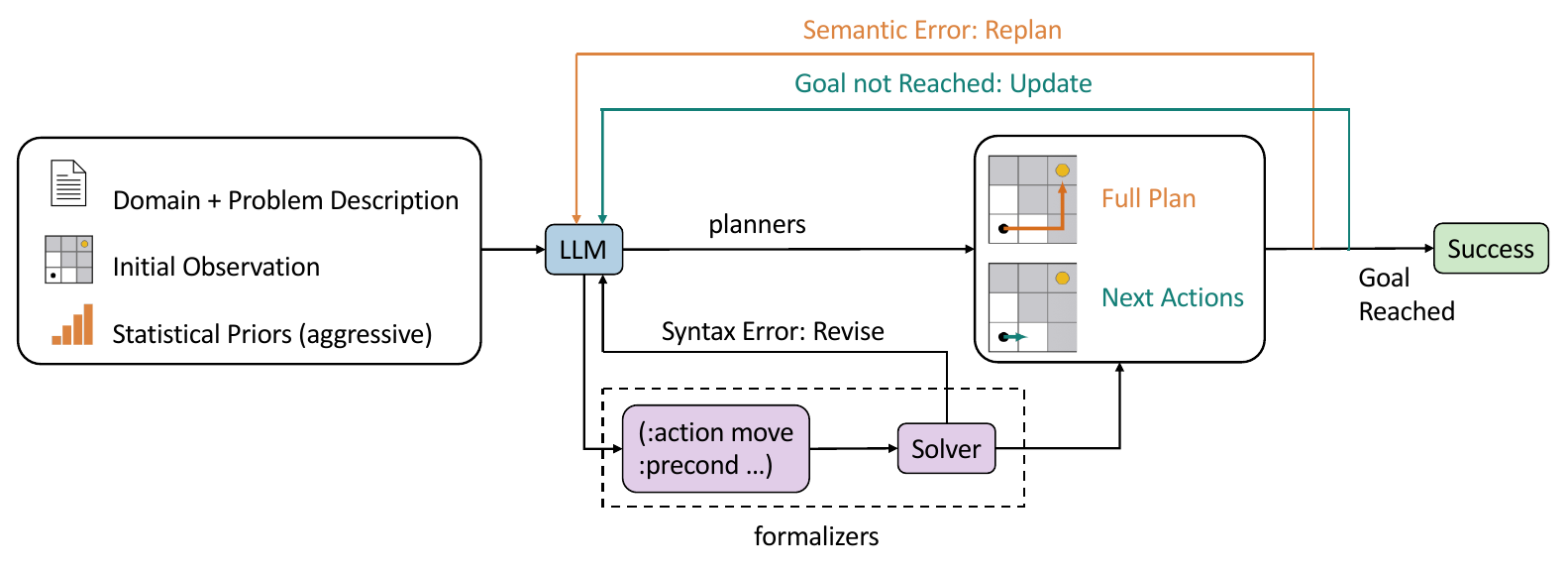}
    \caption{Overview of all planning and formalization methodologies. While conservative methods (dark green) plan using only observations and subgoals, aggressive methods (orange) come up with the entire plan by assuming missing information. All approaches are evaluated using a simulator.}
    \label{fig:pipelines}
\end{figure*}

\paragraph{Planning tasks.}
We first define the task of text-based planning. A planning problem is a tuple $\langle \mathcal{S}, \mathcal{A}, \mathcal{T},
s_0, g\rangle$, where $\mathcal{S}$ is the set of world states,
$\mathcal{A}$ is the set of actions the agent can execute,
$\mathcal{T}:\mathcal{S}\times\mathcal{A}\rightarrow\mathcal{S}$ is a
deterministic transition function, $s_0\in\mathcal{S}$ is the initial state,
and $g$ is a goal condition. The task is to output a sequence of actions
$\pi=[a_0,\dots,a_m]$ such that the state reached by applying $\pi$ from
$s_0$ under $\mathcal{T}$ satisfies $g$. The model never sees this tuple
directly. Instead, it is given a natural language \emph{domain description}
(entity types and what each action does), a \emph{problem description} (the
objects, $g$, and the known facts about $s_0$), and the signatures of the
actions in $\mathcal{A}$. In Coin Collector, for example, $g$ is to hold the
coin located in a named room, and $\mathcal{A}$ consists of \texttt{move}
and \texttt{take}. Optionally, the model also receives a set of
\emph{statistical priors} $K$ about the hidden facts, stated in natural
language (e.g., \emph{86\% of the time, room\_2 is connected directly to the
east of room\_1}; Section~\ref{sec:data}) as a substitute for the common sense the domain lacks.
 
\paragraph{Partial observability.}
Under full observability, the problem description states every fact needed
to plan (e.g., \emph{To the south of room\_1 is room\_9}). Under partial
observability, some facts about $s_0$ are simply not given. For example in Coin Collector, the model is
told the current room and where the coin is located, but none of the connections between the rooms. A missing fact is revealed
only when the agent acts in the environment; after each action it receives
a text observation of its new state, or an error message if the action
could not be executed. 
 
\paragraph{Interaction and success.}
The agent solves a task over a sequence of planning rounds. In round $k$,
the model receives the task and the interaction history $h_k$ (all actions
taken and observations received so far, including failed attempts) and
produces a plan segment $\pi_k$. The segment is executed until it ends or an
action is rejected, and the resulting observations are appended to $h_k$.
A task is solved if the goal is reached within a certain number of rounds;
correctness is judged only by executing in the simulator, never by
comparison to a reference plan, so a successful run may include rejected
actions followed by replanning. 
 
\subsection{Planning Methodologies}
\label{sec:methods}
 
We vary two independent choices, summarized in Figure~\ref{fig:pipelines}.
 
\paragraph{Planner vs.\ formalizer.}
An \textbf{LLM-as-planner} writes the plan segment $\pi_k$ directly as a
sequence of actions. An \textbf{LLM-as-formalizer} instead writes a PDDL
domain and problem file $\Phi_k$ describing what it believes about the
environment, and an off-the-shelf classical planner computes
$\pi_k=\textsc{Solve}(\Phi_k)$. If the PDDL cannot be parsed or solved, the
error is returned to the model for revision.
 
\paragraph{Conservative vs.\ aggressive commitment.}
The two strategies differ only in which facts a plan is allowed to rely on.
A \textbf{conservative} agent relies only on facts stated in the problem
description or observed in $h_k$, and ignores $K$ even when it is given.
Because the goal is usually not reachable from known facts alone, it plans
toward an intermediate subgoal (e.g., moving into an unexplored room), and
extends the plan once the segment succeeds. An \textbf{aggressive} agent
fills in every missing fact with an assumed value---drawn from its own
background knowledge, or from $K$ when provided---and plans all the way to
the goal; it replans only when an action is rejected. In both cases,
replanning continues from the current state rather than restarting. Under this definition, any use of unobserved information, including choosing which room to explore based on a belief about where it leads, constitutes commitment. Conservative agents are therefore instructed to act only on observed facts and are not given $K$, although their choices among observed options may still be implicitly shaped by the LLM’s knowledge.

Each combination corresponds to a family of existing approaches (Section \ref{sec:related}). A formal statement of the execution
semantics as well as definitions for all planning methodologies is given in Appendix~\ref{app: formal}.

\section{Data} \label{sec:data}

We construct benchmarks based on two widely used planning domains to evaluate aggressive and conservative methods under partial observability:

\noindent\textbf{Coin Collector} \cite{yuan2018counting} is a domain to navigate a 2D grid of rooms in order to find a coin. It involves two actions: \textit{move} and \textit{take}.
The domain has no inherent common sense, so the priors serve as a controlled proxy for it. In order to control the statistical priors of the dataset, we create our own problems and environment simulator. To generate our dataset, we over-generate floor plans of 9 rooms, and select a subset of problems where 10 of the room pair statistics are considered significant. A room pair statistic is the number of times that a pair of rooms appear together in a given direction. It is considered significant if it is at least 80\% or at most 20\%. We then select a subset of problems such that 10 room pair statistics over the resulting dataset are significant. All significant statistics are given to the model as priors and accurately reflect the dataset; we do not consider settings where the provided priors are inaccurate. We also evaluate a no-prior setting in which these statistics are withheld. The problems are then used to create natural language problem descriptions. All problem information is given in the fully observable setup, which serves as a control group, while information on room connections is omitted for partially observable problems. To make partially observable problem descriptions, we remove information on \textbf{all} connections in the floor plan. In total we have 100 problems, each consisting of a fully observable and partially observable description. Data examples can be found in Appendix Section \ref{sec:data_examples}.

\noindent\textbf{ALFWorld household tasks.}
ALFWorld \cite{shridhar2021alfworld} is a widely used domain for robotic household tasks. Each game contains several household objects that agents can interact with in either a 3D environment or a natural language environment. To simplify the domain and better support our experiments, we created our own controlled ALFWorld dataset and simulator with natural language interaction only. In each problem, the agent must move a specified object from its initial receptacle to a target receptacle. It involves the following actions: navigate between rooms, inspect receptacles, open and close containers, pick up an object, and place it in or on another receptacle.
To generate our dataset, we build a pipeline based on the original ALFWorld environment and generated 100 source worlds. Each final problem contains 5 rooms and 3 complete receptacle-object pairs in every room. The source and target receptacles required by the goal are always kept, and every room contains at least one surface and one openable container. We then assign every problem a unique connected five-room tree. In total, every problem has 5 rooms, 15 receptacles, 15 movable objects, and one cross-room placement goal. ALFWorld consists of only a partially observable dataset where room connectivity and room contents are omitted. We do not provide statistical priors for this domain, rather the problem information was generated to imitate realistic, commonsense, room contents so that the model must use the knowledge in its training data to solve the problem. Because these realistic room contents inherently carry commonsense priors, ALFWorld has no no-prior setting. More details and a data example can be found in Appendix Section~\ref{sec:alfworld_data} and Appendix Figure \ref{fig: data examples}.

\section{Experimental Setup}

We evaluate the four methods of Section~\ref{sec:methods}. All methods receive the same domain description
and initial observation. On Coin Collector, aggressive methods are run with
and without the statistical priors. Conservative methods never receive them,
because using priors to guide any decision, even which observed room to
explore next, would itself be a form of commitment. A conservative agent with
priors would represent an intermediate level of commitment, which we leave to
future work. Our with-priors comparison therefore contrasts the two ends of
the commitment spectrum rather than isolating prior access. ALFWorld provides
no priors.
 
Under full observability, each method has one planning attempt. Under partial
observability, the attempt limit is the worst case a logical agent could
need: $k+1$ attempts on Coin Collector, where $k$ is the number of hidden
connections, and 20 on ALFWorld (Appendix Section~\ref{app:computation}). Formalizers may
revise PDDL syntax errors up to 3 times, and plans are computed with
\texttt{metric-FF}. Plans are executed in our own simulators
(Appendix Section~\ref{app:lenient}).
 
We report planning accuracy (the percentage of problems solved) and, averaged
per problem, the number of failed attempts, API calls, and output tokens. An
attempt is a planning round that ends in success or a rejected action. API
calls count every LLM call, including conservative plan extensions and PDDL
revisions, and serve as our primary efficiency cost measure (Appendix Section ~\ref{app:computation}).
We evaluate Gemini-3-Flash and DeepSeek-Flash through their APIs and
Qwen3.8-27B with vLLM on one NVIDIA H100 GPU (Appendix Section ~\ref{app:computation}).
Prompts are in Appendix Section~\ref{app:prompts}.

\section{Results}
\label{sec:results}


\begin{figure*}[!t]
    \centering
    \includegraphics[width=\linewidth]{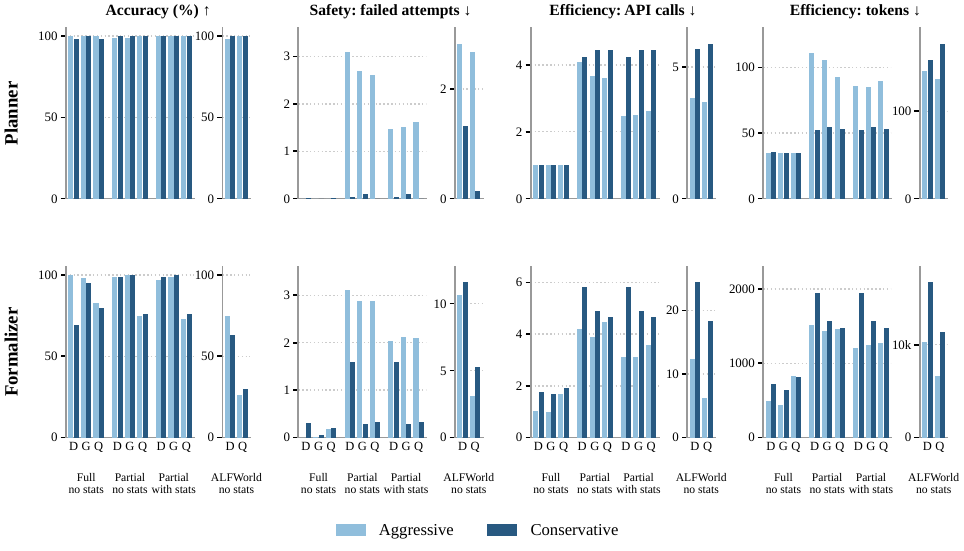} 
    \caption{Results on Coin Collector and ALFWorld for Aggressive vs.\ Conservative Planner and Formalizer using Planning Accuracy, Average Number of Failed Attempts, Average Number of API Calls and Average Output Tokens on DeepSeek-Flash (\texttt{D}), Gemini-3-Flash (\texttt{G}) and Qwen3.8-27B (\texttt{Q}). For ALFWorld, Gemini-3-Flash is not included due to budget constraints.}
    \label{fig:all_results}
\end{figure*}

\paragraph{Conservative vs. Aggressive Methods}

Figure~\ref{fig:all_results} displays the planning accuracy, average number of failed attempts, average number of API Calls and average number of output tokens for Coin Collector across all four methods and three models. It is evident that in most cases \textbf{the performance of aggressive and conservative methods are identical in fully observable environments}. This is expected as the two methods differ in how they handle uncertainty, and full observability removes the uncertainty entirely, reducing both to the same behavior. The remaining variations in behavior stem from prompt differences across methodologies and model randomness. The performance drop for Conservative Formalizer comes from not updating the agent’s location in its PDDL (Appendix \ref{app:formalizer-errors}), a state-tracking error rather than an effect of commitment.

\begin{figure*}[!t]
    \centering
    \includegraphics[width=\linewidth]{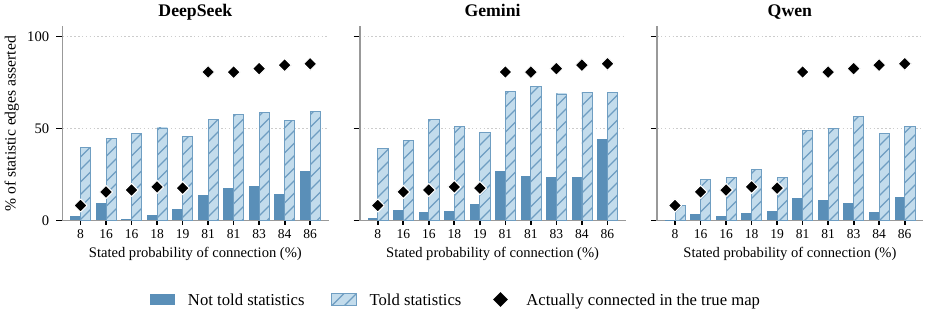}
    \caption{Percentage of the times \texttt{connected} lines in problem files do not appear from observations in the environment, but are given in the statistical priors, for Aggressive Formalizer on Coin Collector.}
    \label{fig: statistics}
\end{figure*}

When uncertainty is introduced, \textbf{conservative methods are safer and aggressive methods never require more API calls than conservative methods, when no statistical priors are introduced.} However, the savings are modest while failed attempts are far higher. Because conservative methods do not assume any information and use only information from the environment, the average number of failed attempts is minimal, but the model will need many API calls to solve the problem. Meanwhile, because aggressive methods must output the entire plan, the number of API calls is the same as the total number of planning attempts needed to solve the problem. The number of failed attempts is much larger. Moreover, because the model is not given any statistical priors, the aggressive methods must blindly assume information that is usually incorrect. Therefore the number of API calls for aggressive methods is similar to their conservative counterparts. When statistical priors are introduced, \textbf{aggressive methods are much more efficient than conservative methods}. Specifically, the average number of API Calls reduces by around 40\% in comparison to conservative methods. Appendix Figures \ref{fig: planner example} and \ref{fig: formalizer example} display illustrative examples of Aggressive and Conservative, Planner and Formalizer on the same example, comparing the number of attempts and API Calls.

In regards to output token count, {the more cost-saving method depends on the output type}. Because the aggressive planner methods needs to output the entire plan, the average number of output tokens is always greater than its conservative counterpart. Meanwhile, the output tokens for conservative formalizer always exceeds the number for aggressive formalizer. 

\paragraph{Partial Observability under Commonsense Environments}

Because we artificially introduce common sense in Coin Collector using statistical priors, we now explore the performance of aggressive and conservative methods on a truly commonsense environment. From Figure \ref{fig:all_results} we see that \textbf{performance findings are upheld in a more commonsense environment}. Conservative Planner had fewer failed attempts than Aggressive Planner and Aggressive Planner took fewer API Calls on average. There are many more failed attempts for Formalizer methods, most likely due to the difficulty of the domain. Moreover, unlike Coin Collector, the average number of failed attempts for Aggressive Formalizer is lower than Conservative Formalizer and Aggressive Formalizer is much more efficient than its conservative counterpart. For token count, conservative methods took more output tokens on average than aggressive methods. The increased efficiency for aggressive methods suggests that the models used the common sense encoded in its training data to make correct assumptions about the data and solve the problems faster than conservative methods.

\paragraph{Are conservative methods actually conservative?}

We next seek to answer the question whether conservative methods actually only use the information given from the environment. From Appendix Figure \ref{fig: conservative methods} we see that \textbf{conservative methods plan and formalize using only information from the environment.} All models and conservative methods used close to zero connections that were not observed from the environment, while connections that are unobserved can be seen in more than half the plans generated by aggressive planner and at least 30\% of connections encoded in PDDL for aggressive formalizer.

\paragraph{How do aggressive formalizers treat statistical priors?}

Since the provided statistical priors for Coin Collector were curated to include different percentages of connections, this raises the question whether models treat statements of high percentage different from statements of low percentage. To answer this, we count the number of times a connection appears in a problem file for PDDL that was not seen in environment observations, but found in the statistical priors given to the model. From Figure \ref{fig: statistics} it is evident that \textbf{aggressive formalizers over-rely on statistical priors.} The percentage of unobserved connections coming from statistical priors increases when the priors are provided to all three models, and statistical priors of higher probability are much more relied upon.

The over-reliance on statistical priors is consistent with the model preferring provided information over using its training data and performing its own reasoning to solve a problem. While the statistical priors given to the model in this scenario were correct, if a user provides incorrect information to models, this can lead to detrimental consequences when applied to real-world problems.

\paragraph{Method Tradeoff}

\begin{figure*}[!t]
\centering
\includegraphics[width=\linewidth]{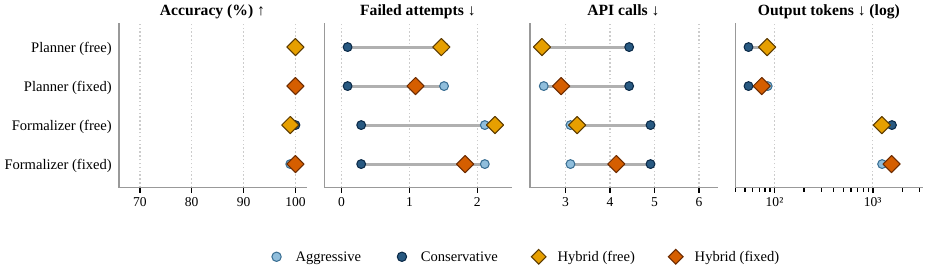}
\caption{Tradeoff on all evaluation metrics across all methods on partially observable Coin Collector data. Only Gemini results are shown, for all models see Appendix Figure \ref{fig: tradeoff}.}
\end{figure*}

From previous sections, it is evident that conservative methods are safer in terms of failed attempts at the cost of efficiency. Meanwhile, aggressive methods have many more failed attempts before solving the problem, but can utilize commonsense knowledge to increase efficiency. This naturally prompts the question whether we can find a middle ground between the safety and efficiency benefits of each method.

To answer this, we explore two types of hybrid reasoning methods for both Planner and Formalizer: free, where the model chooses between reasoning conservatively or aggressively, and fixed where model starts with aggressive reasoning and then turns conservative after a certain number of failed attempts.

We evaluated the performance of these hybrid methods on partially observable Coin Collector problems. From Appendix Figure \ref{fig: tradeoff}, \textbf{hybrid methods provide a compromise between aggressive and conservative methods}. Specifically, Hybrid Planner (fixed) reduces the number of failed attempts and has close to the same API Calls as Aggressive Planner. Gemini-3-Flash and Qwen3.8-27B as Hybrid Formalizer (fixed) also see a good middle ground between the safety of conservative methods and efficiency of Aggressive methods.

The cases where a hybrid method has more failed attempts than both aggressive and conservative methods are DeepSeek-Flash as hybrid formalizer (fixed) and Gemini-3-Flash as hybrid formalizer (free). Both methods perform closely to Aggressive Formalizer, but are worse than Conservative Formalizer. For Gemini-3-Flash, its performance and error types are the same as Aggressive Formalizer, including hallucinated connections that come from statistical priors. Meanwhile for DeepSeek-Flash, the main issue is the model is not tracking the current room of the agent correctly (Appendix \ref{app:formalizer-errors}). While this issue appeared in Conservative Formalizer, this rate of incorrect start rooms for Hybrid Formalizer (fixed) is much higher.

In general, the right level of commitment depends on two costs: the cost of an API call and the cost of a failed action in the environment. \textbf{When no priors or commonsense knowledge are available, we recommend conservative commitment.} In this setting, aggressive methods save a modest number of API calls while failing far more often. \textbf{When reliable priors or commonsense knowledge are available, the choice becomes a genuine tradeoff.} Aggressive methods need far fewer API calls at similar accuracy, but still fail more often than conservative methods. If failures are cheap to recover from, our results suggest that aggressive commitment is preferable; web navigation over a known site structure, where a wrong click can be undone, is one such setting. If failures are costly or irreversible, as in automating a lab protocol, the additional API calls of conservative commitment are likely worth the reduction in failed actions. When both costs matter, hybrid methods offer a middle ground: Hybrid Planner (fixed) reduces failed attempts while keeping API calls close to those of Aggressive Planner. Validating these recommendations in such real-world settings remains future work.

\paragraph{Additional Experiments}

Additional experiments including multiple runs and non-LLM baselines for Coin Collector and a small subset of ALFWorld Gemini experiments can be found in Appendix Section \ref{sec: additional experiments}.


\section{Conclusion}
We introduce two controlled, partially observable environments to compare aggressive and conservative LLM planners and formalizers. Both strategies reach similar accuracy but trade off safety against efficiency. When priors or common sense are available, aggressive methods need far fewer API calls; without them, they save little while failing more often. Conservative methods generally fail less often, though not for formalizers on ALFWorld. We therefore recommend aggressive commitment when reliable priors are available and failures are cheap to recover from, conservative commitment when priors are lacking or failures are costly, and hybrid methods when both matter.

\section{Limitations}

This work focuses on two text-based domains, Coin Collector and ALFWorld (under the MIT License), which are both artificial. While we try to make the domains as realistic as possible, performance in this work may not reflect performance on other real-world domains. Moreover, our recommendations are derived from two controlled text-based domains; whether they transfer to real-world settings such as web navigation or lab automation is untested. This opens potential safety risks when using these methodologies in the real world, such as inefficient spending in real world tasks and consequences when errors occur. Nevertheless, this opens future work to explore aggressive methods further and on larger, more realistic domains.

Priors are computed from the evaluation set itself, so they are perfectly calibrated; we do not test miscalibrated or held-out priors. The risk of incorrect priors is extrapolated from the observed over-reliance, not measured directly. Because using priors is itself a form of commitment in our framework, we do not evaluate agents that use priors only to guide exploration. Such intermediate-commitment agents would help separate the value of prior information from the value of committing to it.

One of our evaluation metrics, planning accuracy, may induce false positives where the plan may happen to be correct, but the generated PDDL files do not actually correctly describe the environment that satisfies the constraint, leading to concerns in interpretability and faithfulness. However, to the best of our knowledge there is no feasible alternative. Nevertheless we found no false positives in a 20-sample manual inspection. While there exists methods to evaluate generated PDDL problem files \cite{zuo2024planetarium} or domain files \cite{coulter2022theory} against a ground-truth, fundamentally there exists non-unique ways to formulate the same constraint. 

Output token counts exclude hidden reasoning tokens, which are not reported consistently across providers, so they underestimate the true generation cost.

\bibliography{anthology, custom}

\appendix

\begin{figure*}[!t]
\centering
\includegraphics[width=\linewidth]{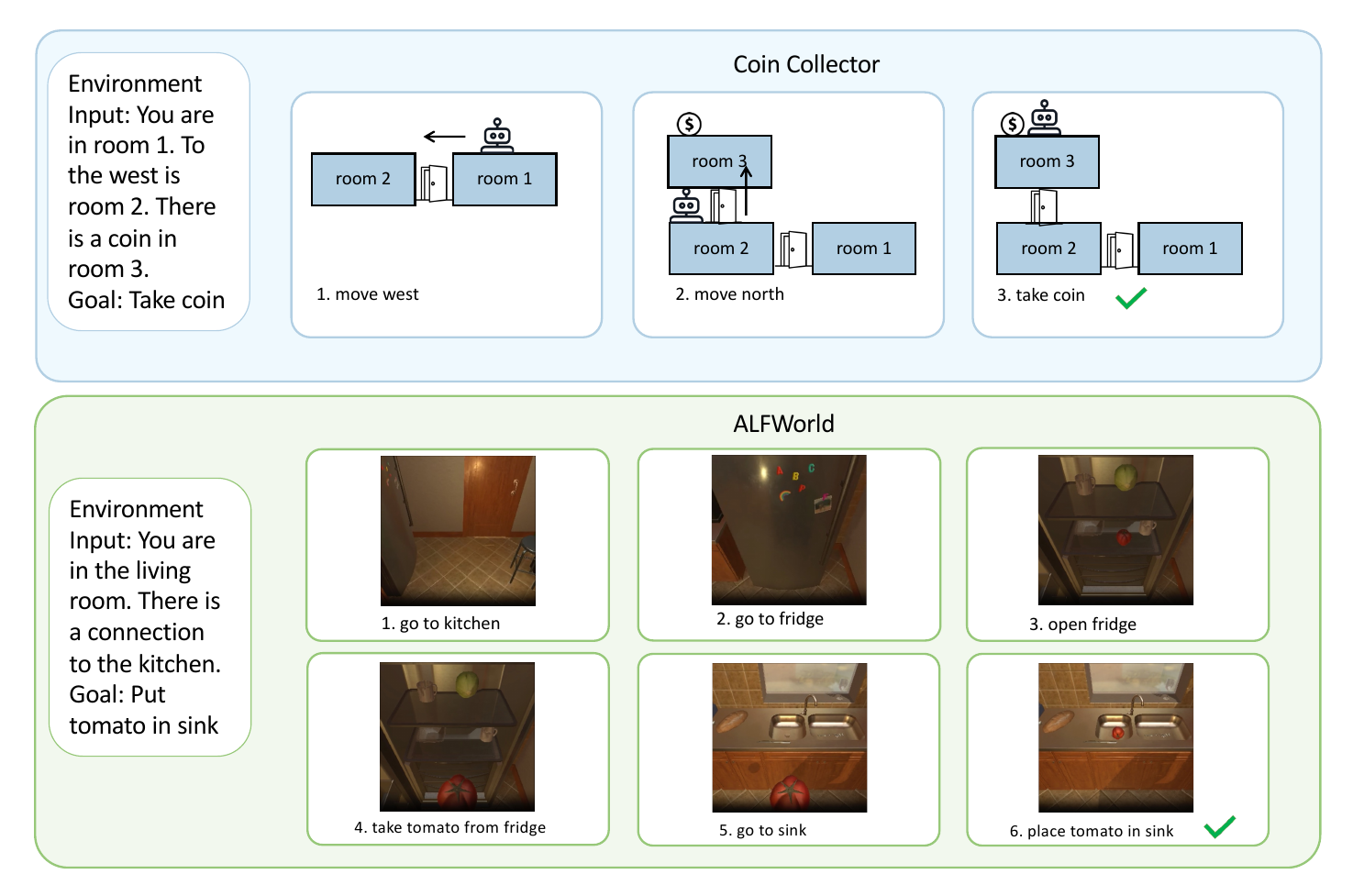}
\caption{Examples of inputs and plans for Coin Collector and ALFWorld. Note that the images provided are for illustration purposes only and our input is purely textual.}
\label{fig: data examples}
\end{figure*}

\begin{figure*}[!t]
\centering
\includegraphics[width=\linewidth]{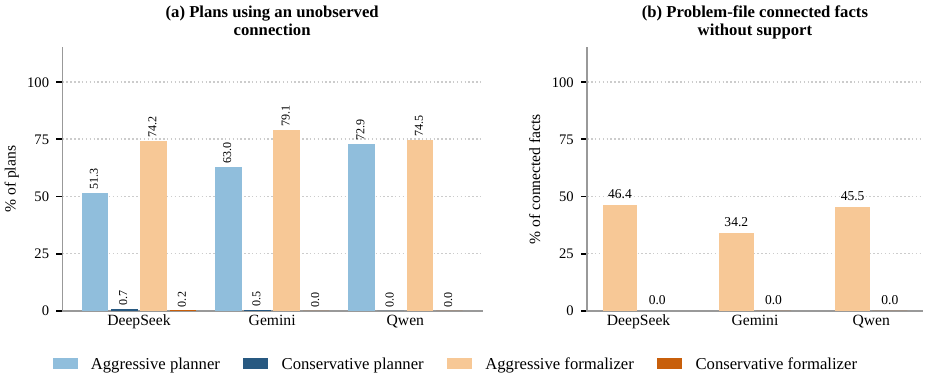}
\caption{(a) The percentage of plans from planner methods that used a connection in Coin Collector that was not observed in the environment. (b) The percentage of connected facts in formalizer methods that contained a connection that did not appear in the environment.}
\label{fig: conservative methods}
\end{figure*}

\begin{figure*}[!t]
    \centering
    \includegraphics[width=\linewidth]{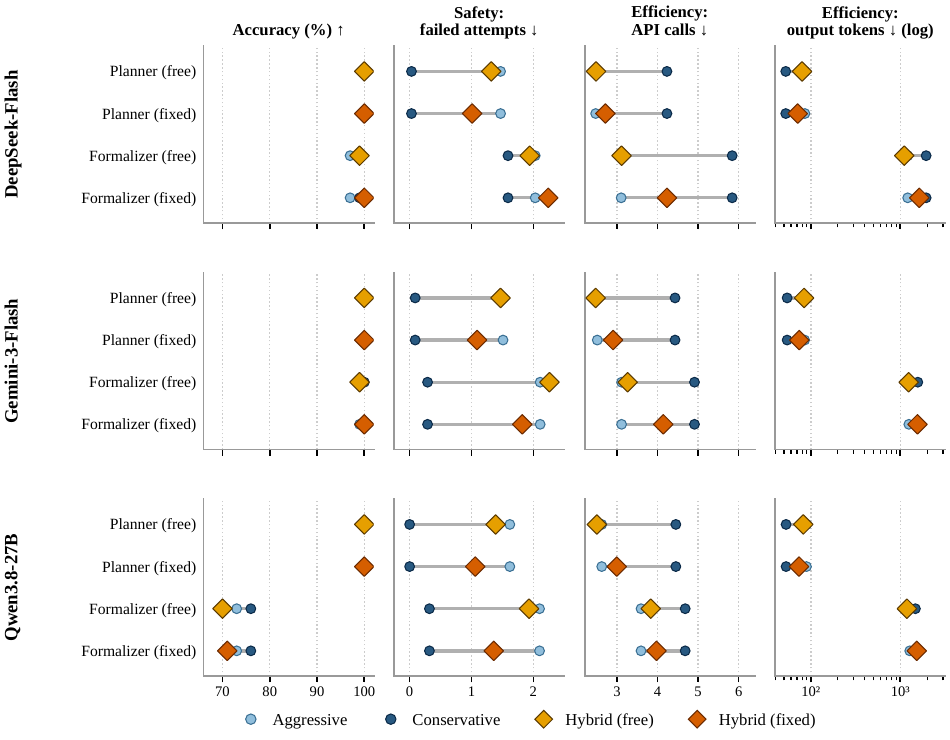}
    \caption{Tradeoff on all evaluation metrics across Conservative, Aggressive and Hybrid (both fixed and free) Planner and Formalizer on partially observable Coin Collector data. Only results when statistics are provided are shown.}
    \label{fig: tradeoff}
\end{figure*}

\section{Formal Task and Methodology Formulation} \label{app: formal}

We formulate interactive text-based planning as decision-making over a discrete-time simulator. Following classical planning paradigms and recent text-based conventions, an agent must reach a target goal state by interacting with an environment. Unlike standard static planning benchmarks, the validity and correctness of a generated plan are determined strictly via online simulator execution rather than offline comparison to ground-truth specification files.

\subsection{Interactive Planning Environment}
\label{subsec:environment}

We define an interactive planning task instance as a quintuple:
\begin{equation}
I = (D_d, D_p^{\rho}, \mathcal{A}', o_0, K)
\end{equation}
where:
\begin{enumerate}[leftmargin=*]
    \item $D_d$ is a natural language domain description detailing relevant entity types, relations, and action semantics.
    \item $D_p^{\rho}$ is the problem description under information setting $\rho \in \{\mathrm{full}, \mathrm{partial}\}$, specifying objects, goal specifications, and initial contextual information.
    \item $\mathcal{A}'$ is the set of action signatures defining valid action names and parameter formats accepted by the simulator.
    \item $o_0 = \mathcal{O}(s_0)$ is the initial console observation generated from the simulator's true underlying state $s_0 \in \mathcal{S}$ via the observation function $\mathcal{O}: \mathcal{S} \rightarrow \Omega$.
    \item $K$ is an optional set of auxiliary common-sense prior statements. When no prior domain knowledge is supplied, $K = \varnothing$.
\end{enumerate}

\paragraph{Observability and Information Settings.} 
The simulator maintains a complete underlying state $s_t \in \mathcal{S}$ at every step $t$, regardless of what the agent observes. The information setting $\rho$ determines how $D_p^{\rho}$ is instantiated:
\begin{itemize}[leftmargin=*]
    \item \textbf{Full Observability ($\rho = \mathrm{full}$):} The model receives $D_p^{\mathrm{full}}$, which contains all facts necessary to construct a complete solution \textit{a priori}.
    \item \textbf{Partial Observability ($\rho = \mathrm{partial}$):} The model receives $D_p^{\mathrm{partial}}$, which omits some facts about $s_0$ (all room connections in Coin Collector; room connections and entity locations in ALFWorld). Omitted facts are revealed only through observations. 
\end{itemize}

\paragraph{Execution Dynamics and Plan Correctness.}
At execution step $t$, given a state $s_t \in \mathcal{S}$ and a grounded action $a_t \in \mathcal{A}$ the simulator evaluates executability via a binary function:
\begin{equation}
\operatorname{Exec}: \mathcal{S} \times \mathcal{A} \rightarrow \{\text{True}, \text{False}\}
\end{equation}
Where $\mathcal{S}$ is the state space and $\mathcal{A}$ is the true action space, each element additionally carrying semantics that $\mathcal{A}'$ omits. If $\operatorname{Exec}(s_t, a_t) = \text{True}$, the environment updates the state as $s_{t+1} = \mathcal{T}(s_t, a_t)$ and yields the observation $o_{t+1} = \mathcal{O}(s_{t+1})$. Otherwise, the state remains unchanged ($s_{t+1} = s_t$), and the simulator returns an execution-error observation $o_{t+1} \in \Omega$ describing why the action was rejected.

The objective is to produce an action sequence $\pi = [a_0, \dots, a_m]$ yielding an execution trajectory $\tau = [o_0, a_0, o_1, a_1, \dots, a_m, o_{m+1}]$. A plan segment, $\pi$, is executable if every action in the segment is accepted
by the simulator. An interaction run is successful if its terminal state
satisfies $s_{m+1} \in \mathcal{S}_g$, where $\mathcal{S}_g \subset \mathcal{S}$ denotes the set of goal states.
A successful run may include rejected actions followed by replanning.

\subsection{Planning and Reasoning Methodologies}
\label{subsec:methodologies}

We index execution steps by $t$ and planning round by $k$; round $k$ begins at step $t_k$. To solve instances given by $I$, we categorize LLM-based planning methodologies along two core axes: the \textit{generation strategy} (LLM-as-Planner vs.\ LLM-as-Formalizer) and the \textit{reasoning paradigm under uncertainty} (Conservative vs.\ Aggressive Reasoning).

\paragraph{Axis 1: Generation Strategy (Planner vs.\ Formalizer).}
\begin{itemize}[leftmargin=*]
    \item \textbf{LLM-as-Planner:} The LLM directly operates as an end-to-end policy. Given the task specification $I$ and interaction trajectory $h_k = [o_0, a_0, \dots, o_{t_{k}}]$, the model directly generates a sequence of executable grounded actions:
    \begin{equation}
    \pi_k = [a_{t_k}, \ldots, a_{t_{k+j}}] = \mathrm{LLM}(I, h_k)
    \end{equation}
    where $j$ is determined by the reasoning paradigm (Axis 2): bounded by verified information under conservative reasoning, or spanning the full remaining plan to the goal under aggressive reasoning.
    \item \textbf{LLM-as-Formalizer:} The LLM acts as a translator that maps the natural language inputs into a formal symbolic representation $\Phi_k$ (e.g., PDDL domain and problem specifications):
    \begin{equation}
    \Phi_k = \mathrm{LLM}(I, h_k)
    \end{equation}
    An external deterministic symbolic solver $\Sigma$ is then invoked to compute a plan that is sound with respect to $\Phi_k$:
    \begin{equation}
    \pi_k = \Sigma(\Phi_k)
    \end{equation}
\end{itemize}

\paragraph{Axis 2: Reasoning Paradigm under Uncertainty (Conservative vs.\ Aggressive).}
Under partial observability ($\rho = \mathrm{partial}$), the initial problem description $D_p^{\mathrm{partial}}$ lacks complete state facts. Models navigate this information gap using two distinct paradigms.

Both paradigms are given the same task $I$, which may or may not include common-sense priors $K$ ($K=\varnothing$ when none are supplied). Conservative reasoning is characterized purely by $h_k$: it commits only to what is verified in the observation history, never drawing on $K$ even when available. Aggressive reasoning instead forms an implicit hypothesis $\hat{s}_k \approx s_k$ over unobserved variables. It draws on the LLM's own parametric priors when $K=\varnothing$ and additionally on supplied common-sense priors when $K \neq \varnothing$, allowing it to commit to the full problem regardless of what $h_k$ supports. This hypothesis is never materialized as a separate output. In both paradigms, if execution fails at some step $t$, the agent replans: $h_k$, now including the record of the failed attempt, is used to generate a new plan for the remaining problem rather than restarting from $a_0$. The full executed trajectory $\pi$ is the concatenation of successive round-level outputs $\pi_k$ across any replanning or extension steps.
\begin{itemize}[leftmargin=*]
    \item \textbf{Conservative Reasoning:} The agent operates strictly over verifiably grounded facts collected in the observation history $h_k$. It refrains from making unverified assumptions about the unobserved state space. Under the LLM-as-Planner strategy, this yields a bounded action sequence extending only as far as grounded facts support, which is a subgoal-directed segment rather than a completed solution. If this segment executes successfully but the goal is not yet reached, the agent extends the plan: $h_k$ is updated with the new observations and a further segment is generated. Under the LLM-as-Formalizer
strategy, this yields a symbolic representation containing only observed
facts, similarly grown incrementally as $h_k$ expands:

\begin{subequations}
\begin{align}
\pi^{\text{cons}}_k &= [a_{t_k}, \ldots, a_{t_k + j}] = \mathrm{LLM}(I \setminus K, h_k) \\
\Phi_k &= \mathrm{LLM}(I \setminus K, h_k)
\end{align}
\end{subequations}

\noindent where $j$ is bounded by the extent of information available in $h_k$.
    \item \textbf{Aggressive Reasoning:}  Unlike the bounded,
evidence-limited commitment above, the agent commits to the full problem regardless of what is currently grounded, substituting hypothesized assumptions for missing facts where $h_k$ is insufficient. Under the LLM-as-Planner
strategy, this yields a complete trajectory to the goal. Under the LLM-as-Formalizer strategy, the external solver $\Sigma$ always
returns a complete plan from $\Phi_k$, so the distinction from the conservative lies in the
informational content of $\Phi_k$, which now contains hypothesized facts alongside observed ones:

\begin{subequations}
\begin{align}
\pi^{\text{agg}}_k &= [a_{t_k}, \ldots, a_m] = \mathrm{LLM}(I, h_k) \\
\Phi_k &= \mathrm{LLM}(I, h_k)
\end{align}
\end{subequations}
\end{itemize}

\section{Additional Computation Information and Experimental Setup} \label{app:computation}
\begin{figure*}[!t]
    \centering
    \includegraphics[width=\linewidth]{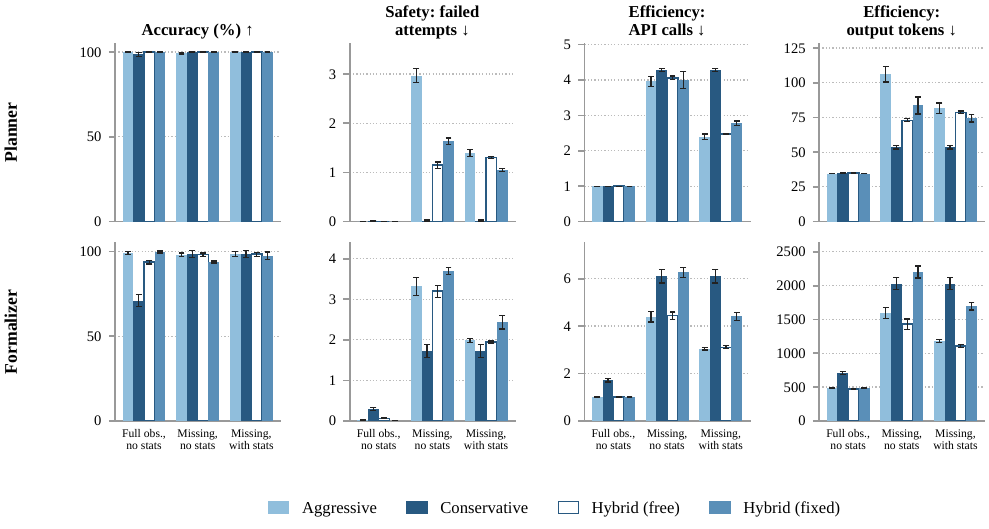}
    \caption{Multiple runs of DeepSeek-Flash on Coin Collector data. Note because Conservative methods were not given statistical priors, results are repeated.}
    \label{fig: multiple runs}
\end{figure*}

\paragraph{Attempt limits.}
The model is given one planning attempt when the problem is fully observable.
Under partial observability, the attempt limit represents the worst case a
logical agent could need given the missing information, rather than a budget
on API calls. For Coin Collector, the model is given $k+1$ attempts, where $k$
is the number of connections omitted from the problem description (8--11 per
problem). For ALFWorld, the model is given
3 receptacles $\times$ 5 rooms $+$ 4 room connections $+$ 1 final attempt
$= 20$ attempts. Formalizers are additionally given up to 3 revision attempts
per output to fix PDDL syntax errors.

\paragraph{Cost counters.}
An \emph{attempt} is a planning round that ends either in success or in a
rejected action; a failed attempt is any attempt that does not reach the goal.
For aggressive methods, each attempt is a single LLM output (a full plan or
PDDL program). For conservative methods, one attempt can contain several LLM
outputs, because the agent extends its plan after every successful segment,
so a new attempt starts only after a rejected action. \emph{API calls} count
every LLM call, including conservative plan extensions and PDDL revision
calls. This is why API calls can exceed attempts even under full
observability. Output tokens are counted as reported by each provider;
reasoning tokens are not exposed consistently across our three models and are
not included. Since every call re-sends the task and the interaction history,
input cost grows with the number of API calls, which we treat as our primary
cost measure.

DeepSeek-Flash was called with default parameters using the default OpenAI API. The following is model information for the model we used:
\begin{itemize}
    \item Model Name: deepseek-flash
\end{itemize}
Gemini-3-Flash was called using the Agent Platform API. The following is information about the model and parameters we used:
\begin{itemize}
    \item Model Name: gemini-3-flash-preview
    \item temperature: 0.4
    \item top\_p: 0.95
    \item max\_output\_tokens: 20,000
\end{itemize}
Qwen3.8-27B was called using vLLM. The following is the information about the model and parameters we used:
\begin{itemize}
    \item Model Name: Qwen/Qwen3.8-27B
    \item max\_model\_len: 60,000
    \item tensor\_parallel\_size: 1
    \item reasoning\_parser: qwen3
    \item temperature: 0.4
    \item top\_p = 0.95
\end{itemize}

\section{Additional Experiments} \label{sec: additional experiments}
\subsection{Multiple Runs}

Because sampling temperature was set to 0.4 for Gemini-3-Flash and Qwen3.8-27B and left at the default for DeepSeek-Flash, model outputs are inherently stochastic. To check that our findings are robust to this randomness, we repeated the Coin Collector experiments three times with DeepSeek-Flash. Figure \ref{fig: multiple runs} shows that, although individual metrics vary across runs, this variation is small relative to the differences between methods, and the qualitative comparisons between methods are consistent across all three runs. Due to time and budget constraints, Gemini-3-Flash and Qwen3.8-27B were run once.

\subsection{ALFWorld Gemini Experiments}
\begin{table}[t]
\centering
\footnotesize
\setlength{\tabcolsep}{3.5pt}
\begin{tabular}{@{}lrrrr@{}}
\toprule
Method & Acc.$\uparrow$ & Fail.$\downarrow$ & API$\downarrow$ & Tok.$\downarrow$ \\
\midrule
Aggr.\ planner & 100 & 4.0 & 5.0 & 178 \\
Cons.\ planner & 100 & 1.5 & 4.7 & 197 \\
\midrule
Aggr.\ formalizer & 90 & 6.1 & 7.5 & 7,603 \\
Cons.\ formalizer & 100 & 5.7 & 18.6 & 16,149 \\
\bottomrule
\end{tabular}
\caption{Gemini-3-Flash on ALFWorld, 10 problems per method. Aggr.\ and Cons.\ denote aggressive and conservative. Acc.\ is accuracy (\%), Fail.\ the average number of failed attempts per problem, API the average number of API calls, and Tok.\ the average number of output tokens}
\label{tab:alfworld-gemini}
\end{table}
Due to budget constraints, we only performed experiments on a 10-problem subset of ALFWorld using Gemini-3-Flash. From Table \ref{tab:alfworld-gemini} we see that performance is similar to that of DeepSeek-Flash and Qwen3.8-27B. The only exception is Conservative Planner is more efficient in terms of API Calls than Aggressive Planner, which may be due to the small number of problems experimented with.

\subsection{Non-LLM Coin Collector Baselines}

We implemented four non-LLM baselines, where three use no priors; the probability-based baseline samples from the dataset-level connection frequencies.

\textbf{Oracle BFS baseline.} 
The oracle breadth-first search (BFS) baseline assumes full observability of the environment. It is given the complete room topology, the direction of every connection, the initial room, and the exact location of the coin. BFS is performed over action-level states. The baseline generates a complete shortest plan before execution and follows it without replanning. Because it has access to information hidden from the other planners, this baseline represents an oracle upper bound rather than a directly comparable partially observable planning method.

\textbf{Frontier BFS baseline.}
The frontier BFS baseline assumes no prior access to the hidden map, coin location, or statistical connection information. It begins with the agent’s initial observation and incrementally constructs a graph containing only observed rooms and connections. At each planning point, BFS finds the shortest route through the discovered graph to the nearest exploration frontier, defined as an observed but unvisited room. The complete route to that frontier is generated and executed before the graph is updated using newly obtained observations. This process continues until the coin is found, no reachable frontier remains, or the action budget is exhausted. Thus, the baseline replans between frontier-reaching routes, rather than after every action.

\textbf{Aggressive BFS baseline.} 
The aggressive BFS baseline receives the partial problem description and initial observation but does not access the hidden environment map or statistical priors. It constructs a hypothetical complete map by preserving observed connections and filling missing connections according to fixed room and direction orderings. BFS then generates a complete plan to the known goal on this hypothetical map before execution begins. The plan is followed unchanged after successful actions and is terminated at the first invalid action. The resulting observations are used to remove contradicted assumptions and construct a revised hypothetical map, from which another complete plan is generated. Although this method is deterministic and reproducible, its unobserved connections are unsupported assumptions and therefore function as deterministic guesses.

In addition, we also implemented a \textbf{probability based baseline}. It uses dataset-level connection frequencies to define a stochastic movement policy. At each step, it samples a direction according to the current room’s probability distribution, removes directions contradicted by environment feedback, and renormalizes the remaining probabilities. It operates without access to the hidden map or coin location and selects actions incrementally rather than generating a complete plan. Multiple random seeds are used to account for sampling variance.

\begin{table}[t]
\centering
\setlength{\tabcolsep}{4pt}
\resizebox{\columnwidth}{!}{%
\begin{tabular}{lcccc}
\toprule
\textbf{Baseline} & \textbf{Att.} & \textbf{Steps} & \textbf{Acc.} & \textbf{1-Att.} \\
\midrule
Oracle BFS     & 1.00 & 4.09  & 100 & 100 \\
Frontier BFS   & 6.09 & 7.15  & 100 & 0   \\
Aggressive BFS & 5.99 & 13.60 & 100 & 3   \\
Probability    & 5.34 & 62.75 & 100 & 53  \\
\bottomrule
\end{tabular}%
}
\caption{Non-LLM baselines on Coin Collector. Att.\ and Steps are the average number of attempts and actions per problem, Acc.\ is planning accuracy (\%), and 1-Att.\ is the percentage of problems solved on the first attempt.}
\label{tab:baselines}
\end{table}

From Table \ref{tab:baselines} we see that besides the Oracle BFS, the non-LLM baselines need more attempts to complete than all LLM methods. 
Finally, aggressive methods when compared to non-LLM baselines are more efficient or comparable to all non-LLM baselines.

\begin{table}[t]
\centering
\setlength{\tabcolsep}{3pt}
\resizebox{\columnwidth}{!}{%
\begin{tabular}{lcccccc}
\toprule
\textbf{Method} & \textbf{Fail.} & \textbf{PDDL} & \textbf{Loc.} & \textbf{Known} & \textbf{Assumed} & \textbf{Rev.} \\
\midrule
\multicolumn{7}{l}{\textit{Full observability}} \\
\multicolumn{7}{l}{\quad DeepSeek-Flash} \\
\quad\quad Aggressive & 0.00 & 0.00 & 0.00 & 0.00 & -- & 0.00 \\
\quad\quad Aggressive + priors & 0.06 & 0.01 & 0.00 & 0.05 & -- & 0.01 \\
\quad\quad Conservative & 0.31 & 0.00 & 0.30 & 0.01 & -- & 0.04 \\
\multicolumn{7}{l}{\quad Gemini-3-Flash} \\
\quad\quad Aggressive & 0.02 & 0.02 & 0.00 & 0.00 & -- & 0.00 \\
\quad\quad Aggressive + priors & 0.09 & 0.03 & 0.00 & 0.06 & -- & 0.00 \\
\quad\quad Conservative & 0.05 & 0.05 & 0.00 & 0.00 & -- & 0.06 \\
\multicolumn{7}{l}{\quad Qwen3.8-27B} \\
\quad\quad Aggressive & 0.17 & 0.17 & 0.00 & 0.00 & -- & 0.83 \\
\quad\quad Aggressive + priors & 0.17 & 0.16 & 0.00 & 0.01 & -- & 0.85 \\
\quad\quad Conservative & 0.20 & 0.19 & 0.01 & 0.00 & -- & 0.88 \\
\midrule
\multicolumn{7}{l}{\textit{Missing all connections}} \\
\multicolumn{7}{l}{\quad DeepSeek-Flash} \\
\quad\quad Aggressive & 3.11 & 0.04 & 0.29 & 0.67 & 2.11 & 0.10 \\
\quad\quad Aggressive + priors & 2.03 & 0.01 & 0.19 & 0.60 & 1.23 & 0.10 \\
\quad\quad Conservative & 1.59 & 0.06 & 1.52 & 0.01 & 0.00 & 0.12 \\
\multicolumn{7}{l}{\quad Gemini-3-Flash} \\
\quad\quad Aggressive & 2.89 & 0.03 & 0.01 & 0.47 & 2.38 & 0.01 \\
\quad\quad Aggressive + priors & 2.11 & 0.01 & 0.02 & 0.79 & 1.29 & 0.01 \\
\quad\quad Conservative & 0.29 & 0.12 & 0.15 & 0.02 & 0.00 & 0.51 \\
\multicolumn{7}{l}{\quad Qwen3.8-27B} \\
\quad\quad Aggressive & 2.88 & 0.25 & 0.47 & 0.49 & 1.67 & 1.06 \\
\quad\quad Aggressive + priors & 2.10 & 0.21 & 0.48 & 0.47 & 0.94 & 0.97 \\
\quad\quad Conservative & 0.32 & 0.25 & 0.07 & 0.00 & 0.00 & 1.22 \\
\bottomrule
\end{tabular}%
}
\caption{Causes of failed attempts for formalizers on Coin Collector (run 1),
averaged per problem. Fail.\ is the total and equals the sum of PDDL (no
solvable PDDL after revisions, or an unparsable action), Loc.\ (wrong agent
location in the problem file), Known (contradicts a given or observed
connection), and Assumed (relies on an unobserved connection). Rev.\ is the
average number of PDDL revision calls per problem.}
\label{tab:formalizer-errors}
\end{table}

\begin{figure}[t]
\centering
\includegraphics[width=\columnwidth]{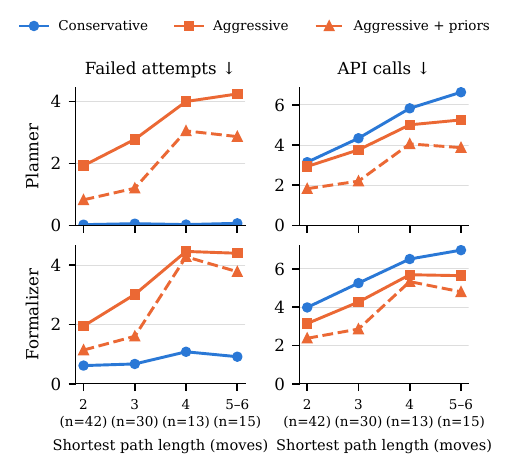}
\caption{Failed attempts and API calls by shortest start-to-coin path length
on Coin Collector without connections, averaged over DeepSeek-Flash,
Gemini-3-Flash, and Qwen3.8-27B (run 1). $n$ is the number of problems.}
\label{fig:path-length}
\end{figure}

\section{Coin Collector Data}
\label{sec:data_examples}

We create our own Coin Collector simulation environment and create problems where the statistical priors regarding the dataset are significant. Here we include an example domain and problem descriptions for this environment, as well as the priors and action heads of the dataset. Listing \ref{lst:coin_collector_actions} is the action heads for the Coin Collector domain and Listing \ref{lst:coin_collector_statistics} is the statistical priors of the dataset. Listings \ref{lst:coin_collector_dd}, \ref{lst:coin_collector_fo_pd} and \ref{lst:coin_collector_po_pf} are domain description, fully observable problem description and partially observable problem description respectively.

\section{ALFWorld Household Data}
\label{sec:alfworld_data}

To create the household dataset, we downsample 100 larger source worlds. For every problem, we keep the rooms containing the source and target receptacles and the agent's initial room. The remaining room slots are selected from the same source world. Within every selected room, we keep three complete receptacle-object pairs, including the source and target receptacles required by the goal. The remaining pairs are selected so that every room has at least one surface and one openable container. We then assign every problem a distinct labeled five-room tree and use the four edges as the room connectivity graph. All goals require moving an object between rooms.

To make partially observable problem descriptions, we remove all four room connections and all object and receptacle locations. An observation only reveals the receptacles in the current room and the names of adjacent rooms. The agent may need to approach and open a container to observe its contents. The maximum of 20 planning attempts consists of 1 initial attempt, 15 possible receptacle inspections, and 4 hidden room connections. We validate that every room graph is a connected tree, all goal entities and initial rooms are kept, and every room contains exactly three receptacles and three objects. In total, all 100 worlds, natural language goals, and labeled tree topologies are unique. The problems span 51 room sets and 60 source-target room pairs.

Listings \ref{lst:aw_actions} and \ref{lst:aw_dd} are the ALFWorld action heads and domain description provided to the model. Listing~\ref{lst:alfworld_partial_problem} is an example of a partially observable problem description. It contains the available entities and the task, while the room graph and entity locations are omitted. 

\begin{table*}[!t]
\centering
\small
\setlength{\tabcolsep}{4pt}
\begin{tabular}{l ccc ccc ccc ccc}
\toprule
& \multicolumn{3}{c}{\textbf{Length 2} ($n{=}42$)} & \multicolumn{3}{c}{\textbf{Length 3} ($n{=}30$)} & \multicolumn{3}{c}{\textbf{Length 4} ($n{=}13$)} & \multicolumn{3}{c}{\textbf{Length 5--6} ($n{=}15$)} \\
\cmidrule(lr){2-4}\cmidrule(lr){5-7}\cmidrule(lr){8-10}\cmidrule(lr){11-13}
\textbf{Method} & Acc. & Fail. & API & Acc. & Fail. & API & Acc. & Fail. & API & Acc. & Fail. & API \\
\midrule
Conservative planner & 100 & 0.02 & 3.14 & 100 & 0.06 & 4.33 & 100 & 0.03 & 5.82 & 100 & 0.07 & 6.62 \\
Aggressive planner & 100 & 1.93 & 2.93 & 98 & 2.78 & 3.76 & 100 & 4.00 & 5.00 & 100 & 4.24 & 5.24 \\
\quad + priors & 100 & 0.83 & 1.83 & 100 & 1.20 & 2.20 & 100 & 3.05 & 4.05 & 100 & 2.87 & 3.87 \\
Conservative formalizer & 91 & 0.61 & 3.98 & 94 & 0.67 & 5.26 & 90 & 1.08 & 6.51 & 89 & 0.91 & 6.98 \\
Aggressive formalizer & 92 & 1.94 & 3.14 & 94 & 3.01 & 4.27 & 82 & 4.46 & 5.69 & 91 & 4.40 & 5.64 \\
\quad + priors & 90 & 1.13 & 2.37 & 89 & 1.60 & 2.86 & 87 & 4.28 & 5.33 & 93 & 3.78 & 4.80 \\
\bottomrule
\end{tabular}
\caption{Coin Collector results without connections by shortest path length,
averaged over the three models (run 1). Acc.\ is planning accuracy (\%),
Fail.\ the average number of failed attempts, and API the average number of
API calls.}
\label{tab:path-length}
\end{table*}

\section{Prompts}
\label{app:prompts}

Listings \ref{lst:prompt_aggressive_planner}, \ref{lst:prompt_aggressive_formalizer}, \ref{lst:prompt_conservative_planner}, \ref{lst:prompt_conservative_formalizer} \ref{lst:hybrid_planner_free}, \ref{lst:hybrid_planner_fixed}, \ref{lst:hybrid_formalizer_free} and \ref{lst:hybrid_formalizer_fixed} are the prompts for the Aggressive Planner, Aggressive Formalizer, Conservative Planner, Conservative Formalizer, Hybrid Planner (free), Hybrid Planner (fixed), Hybrid Formalizer (free) and Hybrid Formalizer (fixed) methods respectively. Statistical priors are presented to the model under the header "Common Sense Knowledge".

\section{Additional Analysis}
\subsection{Direction-Based Move Execution}
\label{app:lenient}
In Coin Collector without connections (run 1, all three models), a move by an aggressive method reached a room other than the one named in its plan in 809 of 3{,}953 planning rounds (20\%). In 805 of these (99.5\%), the round still ended in a rejected action later in the same plan, so direction-based execution delays the detection of a wrong assumption by a step rather than hiding it. Only 4 of 1{,}141 successful aggressive rounds (0.4\%) contained such a mismatch. For conservative methods, mismatches occurred in 100 of 2{,}693 plan segments, and only 2 of 575 successful segments contained one.

\subsection{Formalizer Error Analysis}
\label{app:formalizer-errors}
To separate the effect of commitment from errors in producing PDDL, we
classify every failed attempt of the formalizer methods on Coin Collector
(run 1) by its cause (Table~\ref{tab:formalizer-errors}):
\begin{itemize}
  \item \textbf{PDDL}: no solvable PDDL was produced after 3 revisions, or the
  plan contained an action the simulator cannot parse.
  \item \textbf{Location}: the agent's location in the problem file differs
  from its true location, typically because the model copies the initial
  location instead of updating it.
  \item \textbf{Known}: the failing move contradicts a connection that was
  given in the description or already observed.
  \item \textbf{Assumed}: the failing move relies on a connection the agent
  had not yet observed, which is the failure that aggressive commitment
  deliberately risks.
\end{itemize}
The breakdown sums to the failed attempts in Tables~5--7.
 
Three patterns stand out. First, most failures of aggressive formalizers are
wrong assumptions, as intended, and priors reduce these by 42--46\%, while
other causes stay roughly constant. Second, conservative formalizers almost
never fail on assumptions, but DeepSeek-Flash fails mostly by keeping a stale
agent location in its problem file (96--97\% of its failures). This also
explains its lower accuracy under full observability, where a conservative
formalizer has a single attempt and one stale location ends the run; the drop
therefore reflects state tracking, not commitment. Third, Qwen3.8-27B's
formalizer failures under full observability are almost entirely PDDL errors, and it
needs about one revision call per problem, compared with at most 0.12 for
DeepSeek-Flash. Its lower formalizer accuracy is thus a formalization problem
shared by both strategies.
 
Priors also slightly increase contradictions with known facts for
Gemini-3-Flash (0.47 to 0.79 per problem): in some cases the model reasserts a high-probability connection that it had already observed to be absent, which is consistent with the over-reliance on priors.

\subsection{Results by Path Length}

\label{app:path-length}
The shortest path from the start room to the coin room is 2 moves in 42
problems, 3 in 30, 4 in 13, and 5--6 in 15 (mean 3.09). The number of hidden
connections, which sets the attempt limit, ranges from 8 to 11 and does not
vary with path length. Figure~\ref{fig:path-length} and
Table~\ref{tab:path-length} break down the results on Coin Collector without
connections by path length, averaged over the three models (run 1).
 
API calls grow with path length for all methods, and fastest for conservative
methods, which must observe each room on the way. Failed attempts grow with
path length for aggressive methods but stay near zero for conservative
planners. The safety--efficiency tradeoff therefore holds at every path
length, and both sides of it widen on longer paths. For planners, priors
reduce API calls at every length (by 19--41\%). For formalizers, the
savings are largest on short paths and nearly vanish at length 4, where few
problems are available.

\section{Full Results}
Tables \ref{tab:cc-deepseek-run1}, \ref{tab:cc-gemini-run1}, \ref{tab:cc-qwen3.8-run1}, \ref{tab:cc-deepseek-run2}, \ref{tab:cc-deepseek-run3}, and \ref{tab:alfworld-all} display the full results for all experiments performed in the main body of this paper.

\begin{figure*}
    \centering
    \includegraphics[width=\linewidth]{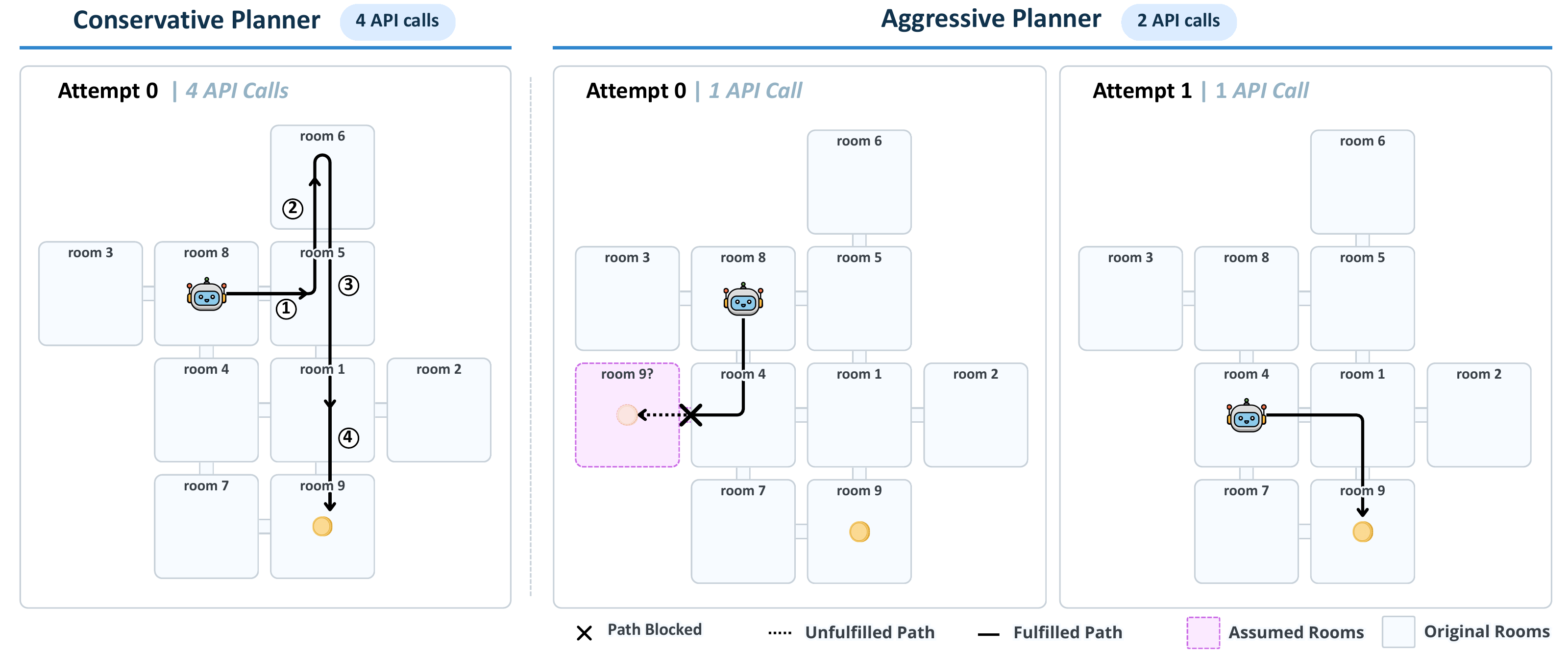}
    \caption{Example of Gemini-3-Flash as Aggressive and Conservative Planner. While Conservative Planner is able to solve the problem in one attempt, Aggressive Planner is able to solve the problem in fewer API Calls.}
    \label{fig: planner example}
\end{figure*}

\begin{figure*}
    \centering
    \includegraphics[width=\linewidth]{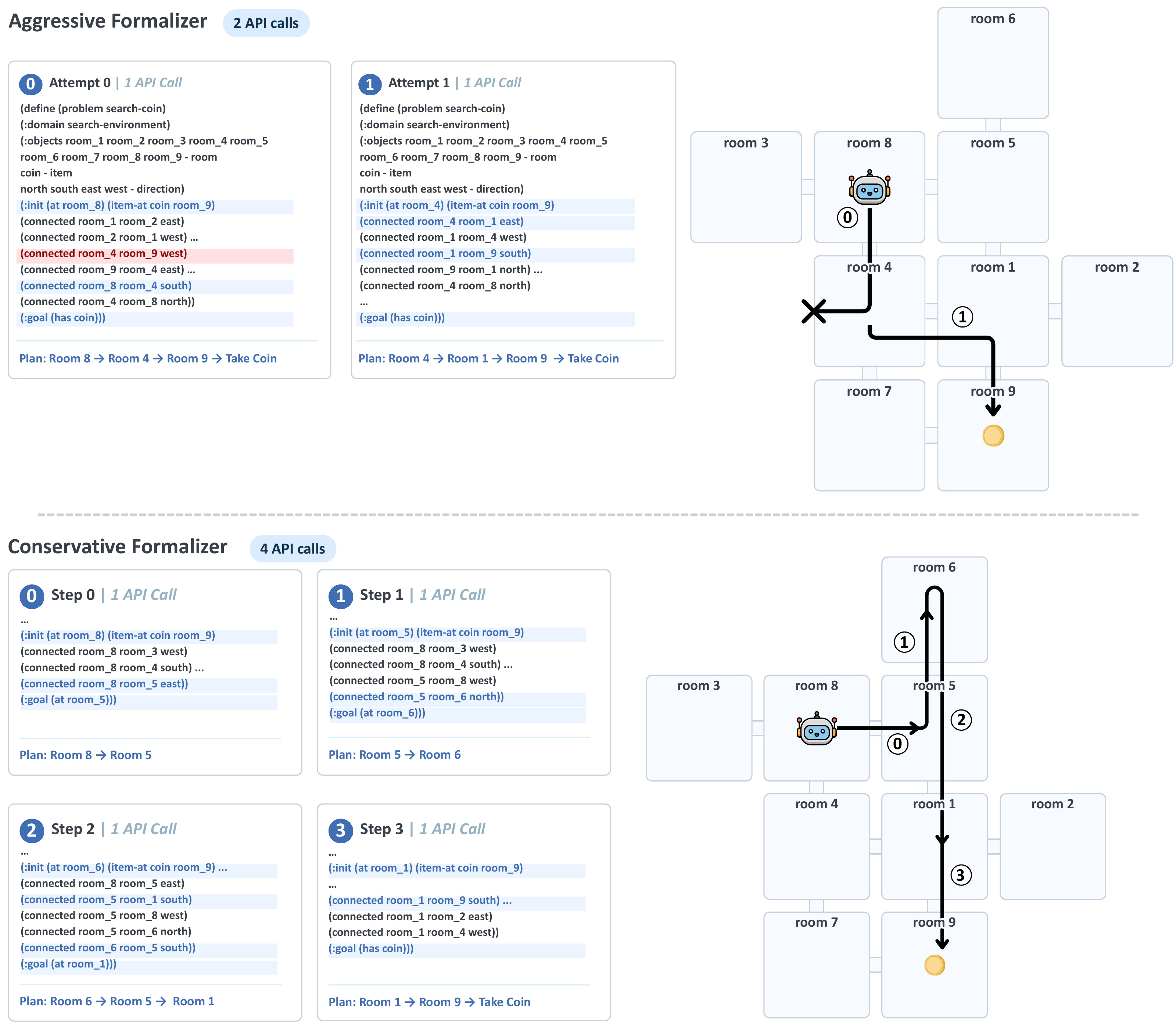}
    \caption{Example of Gemini-3-Flash as Aggressive and Conservative Formalizer. When statistical priors are given under partial observability, the Aggressive Formalizer can leverage these priors and come up with the correct plan with deductive reasoning. }
\label{fig: formalizer example}
\end{figure*}


\clearpage

\onecolumn

\begin{lstlisting}[language=lisp, caption={Statistical Priors for Coin Collector Dataset}, label={lst:coin_collector_statistics}]
86% of the time, room_2 is connected directly to the east of room_1.
16% of the time, room_3 is connected directly to the north of room_2.
84% of the time, room_1 is connected directly to the north of room_9.
16% of the time, room_9 is connected directly to the west of room_4.
81% of the time, room_7 is connected directly to the east of room_6.
19% of the time, room_6 is connected directly to the north of room_5.
81% of the time, room_5 is connected directly to the east of room_8.
18% of the time, room_6 is connected directly to the east of room_4.
83% of the time, room_3 is connected directly to the west of room_8.
8% of the time, room_3 is connected directly to the south of room_5.
    \end{lstlisting}

\begin{lstlisting}[language=lisp, caption={Action Heads for  Coin Collector}, label={lst:coin_collector_actions}]
(move room_1 room_2 direction)
(take item room)
    \end{lstlisting}

\begin{lstlisting}[language=lisp, caption={Action Heads for ALFWorld}, label={lst:aw_actions}]
    ; The text after each semicolon is the complete executable command template.
; Replace only the angle-bracket placeholder with one exact entity name.
; End the command immediately after the entity name. Never append a room qualifier or explanatory text.
(look agent location) ; look
(inventory agent) ; inventory
(examineReceptacle agent receptacle) ; examine <receptacle>
(examineObject agent object) ; examine <object>
(GotoLocation agent from_location to_location receptacle) ; go to <receptacle>
(GoToRoom agent from_location to_location room) ; go to <room>
(OpenObject agent location receptacle) ; open <receptacle>
(CloseObject agent location receptacle) ; close <receptacle>
(PickupObject agent location object receptacle) ; take <object> from <receptacle>
(PutObject agent location object receptacle object_type receptacle_type) ; move <object> to <receptacle>
\end{lstlisting}

\begin{lstlisting}[language=lisp, caption={Domain Description for Coin Collector}, label={lst:coin_collector_dd}]
You are standing in an environment with many rooms. Your task is to search the environment and find the coin. Once you find the coin, take it.

The following actions are allowed: move from one room to another and take an item.
    \end{lstlisting}

\begin{lstlisting}[language=lisp, caption={Domain Description for ALFWorld}, label={lst:aw_dd}]
    You are interacting with a household world composed of rooms, receptacles, and movable objects.
The goal is to place a target object into or onto a target receptacle.

Exact command forms:
- look
- inventory
- examine <object or receptacle>
- go to <receptacle>
- go to <room>
- open <receptacle>
- close <receptacle>
- take <object> from <receptacle>
- move <object> to <receptacle>

Interaction rules:
1. You must first go to a receptacle before interacting with it.
2. Room navigation is direct. "go to <room>" is valid only when that room is listed under "Adjacent rooms" in the latest room observation.
3. Room observations show the current room's receptacles and the names of adjacent rooms. Other rooms' contents are not revealed by this observation.
4. To take an object, you must be at the receptacle containing it. If that receptacle is closed, open it first.
5. You can carry at most one movable object at a time.
6. To move an object to a receptacle, you must be holding the object and be at the target receptacle. If the target receptacle is closed, open it first.
7. Objects inside a closed receptacle may only become visible after you approach and open that receptacle.
8. Use only rooms, receptacles, and objects given in the problem description or revealed by environment observations. Do not invent additional entities.
9. Every command must end immediately after its final entity name. Do not append qualifiers or explanatory text.

Use the problem description, environment observations, and available action heads to choose actions that satisfy these rules.

\end{lstlisting}

    \begin{lstlisting}[language=lisp, caption={Fully Observable Problem Description for Coin Collector}, label={lst:coin_collector_fo_pd}]
There are 9 rooms in this environment: room_1, room_2, room_3, room_4, room_5, room_6, room_7, room_8 and room_9.
You are in room_4.
To the south of room_1 is room_9.
To the south of room_2 is room_6.
To the east of room_3 is room_8.
To the north of room_3 is room_7.
To the west of room_4 is room_9.
To the east of room_5 is room_9.
To the west of room_5 is room_8.
To the east of room_6 is room_7.
There is a coin in room_7.
    \end{lstlisting}

\begin{lstlisting}[language=lisp, caption={Partially Observable Problem Description for Coin Collector}, label={lst:coin_collector_po_pf}]
There are 9 rooms in this environment: room_1, room_2, room_3, room_4, room_5, room_6, room_7, room_8 and room_9.
You are in room_4.

There is a coin in room_7.
    \end{lstlisting}

\begin{lstlisting}[language=lisp, caption={Partially observable household problem description}, label={lst:alfworld_partial_problem}]
The rooms are: living room, bedroom, dining room,
guest room, hallway.
The receptacles are: bar cart, bed, bookcase cabinet,
buffet cabinet, closet, end table, entryway drawer,
guest nightstand, guest nightstand drawer, hall cabinet,
luggage rack, serving cart basket, shoe rack top,
tv stand, wardrobe drawer.
The objects are: jeans, jewelry box, key fob, magazine,
raincoat, reading lamp, scarf, shoe polish, slippers pair,
sugar bowl, tablecloth, teapot, video game case,
water bottle, water glass.
You are in the dining room.
Your task is to put slippers pair in the hall cabinet.
\end{lstlisting}

\begin{lstlisting}[language=lisp, caption={Prompts for Aggressive Planner}, label={lst:prompt_aggressive_planner}]
initial_plan_random = """Given the information below, output the plan that would solve the problem. Note that there might be information missing, in which case you should infer what is missing.

The output should be a list of strings written inside a JSON object in the following format:
{{
  "plan": ...
}}

Domain Description:
{domain_description}

Problem Description:
{problem_description}

Available Actions:
{available_actions}

Initial Observation:
{initial_observation}
"""

initial_plan_statistics = """Given the information below, output the plan that would solve the problem. Note that there might be information missing, in which case you should infer what is missing.

The output should be a list of strings written inside a JSON object in the following format:
{{
  "plan": ...
}}

Domain Description:
{domain_description}

Problem Description:
{problem_description}

Available Actions:
{available_actions} 

Common Sense Knowledge:
{statistics_knowledge}

Initial Observation:
{initial_observation}
"""

re_plan_random = """In the following problem, the generated plan is incorrect either due to incorrect assumptions or other mistakes. Given only the information below, write a new plan, starting from the failed step, that would solve the problem. Do not start from the initial state. Note that there might be information missing, in which case you should infer what is missing.

The output should be a list of strings written inside a JSON object in the following format:
{{
  "plan": ...
}}

Domain Description:
{domain_description}

Problem Description:
{problem_description}

Available Actions:
{available_actions}

Incorrect Generated Plan:
{previous_plan}

Trajectory for Incorrect Plan:
{previous_trajectory}
"""

re_plan_statistics = """In the following problem, the generated plan is incorrect either due to incorrect assumptions or other mistakes. Given only the information below, write a new plan, starting from the failed step, that would solve the problem. Do not start from the initial state. Note that there might be information missing, in which case you should infer what is missing.

The output should be a list of strings written inside a JSON object in the following format:
{{
  "plan": ...
}}

Domain Description:
{domain_description}

Problem Description:
{problem_description}

Available Actions:
{available_actions}

Common Sense Knowledge:
{statistics_knowledge}

Incorrect Generated Plan:
{previous_plan}

Trajectory for Incorrect Plan:
{previous_trajectory}"""
    \end{lstlisting}

\begin{lstlisting}[language=lisp, caption={Prompts for Aggressive Formalizer}, label={lst:prompt_aggressive_formalizer}]
initial_formalize_random = """You are a PDDL expert.

Given the information below, write the domain and problem files in minimal PDDL. Note that there might be information missing, in which case you should infer what is missing.

Return a JSON object in the following format:
{{
  "domain file": ...,
  "problem file":...
}}

Domain Description:
{domain_description}

Problem Description:
{problem_description}

Available Actions:
{available_actions}

Initial Observation:
{initial_observation}
"""

initial_formalize_statistics = """You are a PDDL expert.

Given the information below, write the domain and problem files in minimal PDDL. Note that there might be information missing, in which case you should infer what is missing.

Return a JSON object in the following format:
{{
  "domain file": ...,
  "problem file":...
}}

Domain Description:
{domain_description}

Problem Description:
{problem_description}

Available Actions:
{available_actions}

Common Sense Knowledge:
{statistics_knowledge}

Initial Observation:
{initial_observation}
"""

revision_random = """You are a PDDL expert.

The following example and generated PDDL contains an error. Using only the information below, revise the PDDL to remove the error. Note there might be information missing, in which case you should infer what is missing.

Return a JSON object in the following format:
{{
  "domain file": ...,
  "problem file":...
}}

Domain Description:
{domain_description}

Problem Description:
{problem_description}

Available Actions:
{available_actions}

Generated Domain File:
{previous_domain_file}

Generated Problem File:
{previous_problem_file}

Error Message:
{error_message}
"""

revision_statistics = """You are a PDDL expert.

The following example and generated PDDL contains an error. Using only the information below, revise the PDDL to remove the error. Note there might be information missing, in which case you should infer what is missing.

Return a JSON object in the following format:
{{
  "domain file": ...,
  "problem file":...
}}

Domain Description:
{domain_description}

Problem Description:
{problem_description}

Available Actions:
{available_actions}

Common Sense Knowledge:
{statistics_knowledge}

Generated Domain File:
{previous_domain_file}

Generated Problem File:
{previous_problem_file}

Error Message:
{error_message}
"""

re_formalize_random = """You are a PDDL expert.

In the following problem, the found plan from the domain and problem files is incorrect either due to incorrect assumptions or other mistakes. Given only the information below, write a new domain and problem file in minimal PDDL that would solve the problem, starting from the failed step. Do not start from the initial state. Note there might be information missing, in which case you should infer what is missing.

Return a JSON object in the following format:
{{
  "domain file": ...,
  "problem file":...
}}

Domain Description:
{domain_description}

Problem Description:
{problem_description}

Available Actions:
{available_actions}

Generated Domain File:
{previous_domain_file}

Generated Problem File:
{previous_problem_file}

Incorrect Found Plan:
{previous_plan}

Trajectory for Incorrect Plan:
{previous_trajectory}
"""

re_formalize_statistics = """You are a PDDL expert.

In the following problem, the found plan from the domain and problem files is incorrect either due to incorrect assumptions or other mistakes. Given only the information below, write a new domain and problem file in minimal PDDL that would solve the problem, starting from the failed step. Do not start from the initial state. Note there might be information missing, in which case you should infer what is missing.

Return a JSON object in the following format:
{{
  "domain file": ...,
  "problem file":...
}}

Domain Description:
{domain_description}

Problem Description:
{problem_description}

Available Actions:
{available_actions}

Common Sense Knowledge:
{statistics_knowledge}

Generated Domain File:
{previous_domain_file}

Generated Problem File:
{previous_problem_file}

Incorrect Found Plan:
{previous_plan}

Trajectory for Incorrect Plan:
{previous_trajectory}"""
\end{lstlisting}

\begin{lstlisting}[language=lisp, caption={Prompts for Conservative Planner}, label={lst:prompt_conservative_planner}]
get_action_prompt_json = """You are a Planning expert.

Using ONLY the information below, write an action or a series of actions that would solve the problem. Do NOT assume any information. Note the problem may not be solvable with the given information, in which case you should come up with a tangible subgoal for the problem. The actions must be the same format as the available actions or you will be asked to try again or revert to the beginning.

The output should be a list of strings written inside a JSON object in the following format:
{{
  "plan": ...
}}

Do NOT add extra text.

Domain Description:
{domain_description}

Problem Description:
{problem_description}

Current Trajectory:
{trajectory}

Available Actions:
{available_actions}
"""

re_plan_prompt_json = """You are a Planning expert.

Using ONLY the information below, write an action or a series of actions that would solve the problem, starting from the failed step. Do not start from the initial state. Do NOT assume any information. Note the problem may not be solvable with the given information, in which case you should come up with a tangible subgoal for the problem. The actions must be the same format as the available actions or you will be asked to try again or revert to the beginning.

The output should be a list of strings written inside a JSON object in the following format:
{{
  "plan": ...
}}

Do NOT add extra text.

Domain Description:
{domain_description}

Problem Description:
{problem_description}

Current Trajectory:
{trajectory}

Available Actions:
{available_actions}
"""
\end{lstlisting}

\begin{lstlisting}[language=lisp, caption={Prompts for Conservative Formalizer}, label={lst:prompt_conservative_formalizer}]
initial_observation_prompt_json = """You are a PDDL expert.

Using ONLY the information below, write the domain and problem file in minimal PDDL. Do NOT assume any information. Note the problem may not be solvable with the given information, in which case you should come up with a tangible subgoal for the problem.

Return ONLY a JSON object in the following format:
{{
  "domain file": ...,
  "problem file":...
}}

Do NOT add extra text.

Domain Description:
{domain_description}

Problem Description:
{problem_description}

Trajectory:
{trajectory}

Available Actions:
{available_actions}
"""

revision_prompt_json = """You are a PDDL expert.

The following generated PDDL contains an error. Using ONLY the information below, revise the PDDL to remove the error. Do NOT assume any information. Note the problem may not be solvable with the given information, in which case you should come up with a tangible subgoal for the problem.

Return ONLY a JSON object in the following format:
{{
  "domain file": ...,
  "problem file":...
}}

Do NOT add extra text.

Domain Description:
{domain_description}

Problem Description:
{problem_description}

Trajectory:
{trajectory}

Available Actions:
{available_actions}

Domain File:
{domain_file}

Problem File:
{problem_file}

Error:
{error}
"""

re_generate_prompt_json = """You are a PDDL expert.

In the following problem, the found plan from the domain and problem files is incorrect. Given only the information below, write a new domain and problem file in minimal PDDL that would solve the problem, starting from the failed step. Do not start from the initial state. Do NOT assume any information. Note the problem may not be solvable with the given information, in which case you should come up with a tangible subgoal for the problem.

Return a JSON object in the following format:
{{
  "domain file": ...,
  "problem file":...
}}

Domain Description:
{domain_description}

Problem Description:
{problem_description}

Available Actions:
{available_actions}

Generated Domain File:
{previous_domain_file}

Generated Problem File:
{previous_problem_file}

Incorrect Found Plan:
{previous_plan}

Trajectory for Incorrect Plan:
{previous_trajectory}"""

grow_pddl_prompt_json = """You are a PDDL expert.

Using ONLY the following PDDL files and plan trajectory, modify the PDDL files based on the new observations. Do NOT assume any information. Note the problem may not be solvable with the given information, in which case you should come up with a tangible subgoal for the problem.

Return ONLY a JSON object in the following format:
{{
  "domain file": ...,
  "problem file":...
}}

Do NOT add extra text.

Domain Description:
{domain_description}

Problem Description:
{problem_description}

Available Actions:
{available_actions}

Domain File:
{domain_file}

Problem File:
{problem_file}

Trajectory:
{trajectory}"""



\end{lstlisting}

\begin{lstlisting}[language=lisp, caption={Prompts for Hybrid Planner (Free)}, label={lst:hybrid_planner_free}]
    get_action_prompt = """You are a Planning expert.

Using the information below, write an action or a series of actions that would solve the problem. Note the problem may not be solvable with the given information, in which case you should come up with a tangible subgoal for the problem OR infer what is missing. The actions must be the same format as the available actions or you will be asked to try again or revert to the beginning.

The output should be a list of strings written inside a JSON object in the following format:
{{
  "plan": ...
}}

Do NOT add extra text.

Domain Description:
{domain_description}

Problem Description:
{problem_description}

Current Trajectory:
{trajectory}

Available Actions:
{available_actions}
"""

re_plan_prompt = """You are a Planning expert.

Using the information below, write an action or a series of actions that would solve the problem, starting from the failed step. Do not start from the initial state. Note the problem may not be solvable with the given information, in which case you should come up with a tangible subgoal for the problem OR infer what is missing. The actions must be the same format as the available actions or you will be asked to try again or revert to the beginning.

The output should be a list of strings written inside a JSON object in the following format:
{{
  "plan": ...
}}

Do NOT add extra text.

Domain Description:
{domain_description}

Problem Description:
{problem_description}

Current Trajectory:
{trajectory}

Available Actions:
{available_actions}
"""



get_action_prompt_statistics = """You are a Planning expert.

Using the information below, write an action or a series of actions that would solve the problem. Note the problem may not be solvable with the given information, in which case you should come up with a tangible subgoal for the problem OR infer what is missing. The actions must be the same format as the available actions or you will be asked to try again or revert to the beginning.

The output should be a list of strings written inside a JSON object in the following format:
{{
  "plan": ...
}}

Do NOT add extra text.

Domain Description:
{domain_description}

Problem Description:
{problem_description}

Common Sense Knowledge:
{statistics_knowledge}

Current Trajectory:
{trajectory}

Available Actions:
{available_actions}
"""

re_plan_prompt_statistics = """You are a Planning expert.

Using the information below, write an action or a series of actions that would solve the problem, starting from the failed step. Do not start from the initial state. Note the problem may not be solvable with the given information, in which case you should come up with a tangible subgoal for the problem OR infer what is missing. The actions must be the same format as the available actions or you will be asked to try again or revert to the beginning.

The output should be a list of strings written inside a JSON object in the following format:
{{
  "plan": ...
}}

Do NOT add extra text.

Domain Description:
{domain_description}

Problem Description:
{problem_description}

Common Sense Knowledge:
{statistics_knowledge}

Current Trajectory:
{trajectory}

Available Actions:
{available_actions}
"""
\end{lstlisting}

\begin{lstlisting}[language=lisp, caption={Prompts for Hybrid Planner (fixed)}, label={lst:hybrid_planner_fixed}]
  generate_plan_aggressive_random = """Given the information below, output the plan that would solve the problem. Note that there might be information missing, in which case you should infer what is missing.

The output should be a list of strings written inside a JSON object in the following format:
{{
  "plan": ...
}}

Domain Description:
{domain_description}

Problem Description:
{problem_description}

Available Actions:
{available_actions}

Initial Observation:
{initial_observation}
"""

generate_plan_aggressive_statistics = """Given the information below, output the plan that would solve the problem. Note that there might be information missing, in which case you should infer what is missing.

The output should be a list of strings written inside a JSON object in the following format:
{{
  "plan": ...
}}

Domain Description:
{domain_description}

Problem Description:
{problem_description}

Available Actions:
{available_actions} 

Common Sense Knowledge:
{statistics_knowledge}

Initial Observation:
{initial_observation}
"""

re_plan_aggressive_random = """In the following problem, the generated plan is incorrect either due to incorrect assumptions or other mistakes. Given only the information below, write a new plan, starting from the failed step, that would solve the problem. Do not start from the initial state. Note that there might be information missing, in which case you should infer what is missing.

The output should be a list of strings written inside a JSON object in the following format:
{{
  "plan": ...
}}

Domain Description:
{domain_description}

Problem Description:
{problem_description}

Available Actions:
{available_actions}

Incorrect Generated Plan:
{previous_plan}

Trajectory for Incorrect Plan:
{previous_trajectory}
"""

re_plan_aggressive_statistics = """In the following problem, the generated plan is incorrect either due to incorrect assumptions or other mistakes. Given only the information below, write a new plan, starting from the failed step, that would solve the problem. Do not start from the initial state. Note that there might be information missing, in which case you should infer what is missing.

The output should be a list of strings written inside a JSON object in the following format:
{{
  "plan": ...
}}

Domain Description:
{domain_description}

Problem Description:
{problem_description}

Available Actions:
{available_actions}

Common Sense Knowledge:
{statistics_knowledge}

Incorrect Generated Plan:
{previous_plan}

Trajectory for Incorrect Plan:
{previous_trajectory}
"""

generate_plan_conservative_prompt = """You are a Planning expert.

Using ONLY the information below, write an action or a series of actions that would solve the problem. Do NOT assume any information. Note the problem may not be solvable with the given information, in which case you should come up with a tangible subgoal for the problem. The actions must be the same format as the available actions or you will be asked to try again or revert to the beginning.

The output should be a list of strings written inside a JSON object in the following format:
{{
  "plan": ...
}}

Do NOT add extra text.

Domain Description:
{domain_description}

Problem Description:
{problem_description}

Current Trajectory:
{trajectory}

Available Actions:
{available_actions}
"""

re_plan_conservative_prompt = """
You are a Planning expert.

Using ONLY the information below, write an action or a series of actions that would solve the problem, starting from the failed step. Do not start from the initial state. Do NOT assume any information. Note the problem may not be solvable with the given information, in which case you should come up with a tangible subgoal for the problem. The actions must be the same format as the available actions or you will be asked to try again or revert to the beginning.

The output should be a list of strings written inside a JSON object in the following format:
{{
  "plan": ...
}}

Do NOT add extra text.

Domain Description:
{domain_description}

Problem Description:
{problem_description}

Current Trajectory:
{trajectory}

Available Actions:
{available_actions}
"""  
\end{lstlisting}

\begin{lstlisting}[language=lisp, caption={Prompts for Hybrid Formalizer (free)}, label={lst:hybrid_formalizer_free}]
    initial_observation_prompt = """You are a PDDL expert.

Using the information below, write the domain and problem file in minimal PDDL. Note the problem may not be solvable with the given information, in which case you should come up with a tangible subgoal for the problem OR infer what is missing.

Return ONLY a JSON object in the following format:
{{
  "domain file": ...,
  "problem file":...
}}

Do NOT add extra text.

Domain Description:
{domain_description}

Problem Description:
{problem_description}

Trajectory:
{trajectory}

Available Actions:
{available_actions}
"""

revision_prompt = """You are a PDDL expert.

The following generated PDDL contains an error. Using the information below, revise the PDDL to remove the error. Note the problem may not be solvable with the given information, in which case you should come up with a tangible subgoal for the problem OR what is missing.

Return ONLY a JSON object in the following format:
{{
  "domain file": ...,
  "problem file":...
}}

Do NOT add extra text.

Domain Description:
{domain_description}

Problem Description:
{problem_description}

Trajectory:
{trajectory}

Available Actions:
{available_actions}

Domain File:
{domain_file}

Problem File:
{problem_file}

Error:
{error}
"""

re_generate_prompt = """You are a PDDL expert.

In the following problem, the found plan from the domain and problem files is incorrect. Given the information below, write a new domain and problem file in minimal PDDL that would solve the problem, starting from the failed step. Do not start from the initial state. Note the problem may not be solvable with the given information, in which case you should come up with a tangible subgoal for the problem OR infer what is missing.

Return a JSON object in the following format:
{{
  "domain file": ...,
  "problem file":...
}}

Domain Description:
{domain_description}

Problem Description:
{problem_description}

Available Actions:
{available_actions}

Generated Domain File:
{previous_domain_file}

Generated Problem File:
{previous_problem_file}

Incorrect Found Plan:
{previous_plan}

Trajectory for Incorrect Plan:
{previous_trajectory}"""

grow_pddl_prompt = """You are a PDDL expert.

Using the following PDDL files and plan trajectory, modify the PDDL files based on the new observations. Note the problem may not be solvable with the given information, in which case you should come up with a tangible subgoal for the problem OR infer what is missing.

Return ONLY a JSON object in the following format:
{{
  "domain file": ...,
  "problem file":...
}}

Do NOT add extra text.

Domain Description:
{domain_description}

Problem Description:
{problem_description}

Available Actions:
{available_actions}

Domain File:
{domain_file}

Problem File:
{problem_file}

Trajectory:
{trajectory}"""



initial_observation_prompt_statistics = """You are a PDDL expert.

Using the information below, write the domain and problem file in minimal PDDL. Note the problem may not be solvable with the given information, in which case you should come up with a tangible subgoal for the problem OR infer what is missing.

Return ONLY a JSON object in the following format:
{{
  "domain file": ...,
  "problem file":...
}}

Do NOT add extra text.

Domain Description:
{domain_description}

Problem Description:
{problem_description}

Common Sense Knowledge:
{statistics_knowledge}

Trajectory:
{trajectory}

Available Actions:
{available_actions}
"""

revision_prompt_statistics = """You are a PDDL expert.

The following generated PDDL contains an error. Using the information below, revise the PDDL to remove the error. Note the problem may not be solvable with the given information, in which case you should come up with a tangible subgoal for the problem OR what is missing.

Return ONLY a JSON object in the following format:
{{
  "domain file": ...,
  "problem file":...
}}

Do NOT add extra text.

Domain Description:
{domain_description}

Problem Description:
{problem_description}

Common Sense Knowledge:
{statistics_knowledge}

Trajectory:
{trajectory}

Available Actions:
{available_actions}

Domain File:
{domain_file}

Problem File:
{problem_file}

Error:
{error}
"""

re_generate_prompt_statistics = """You are a PDDL expert.

In the following problem, the found plan from the domain and problem files is incorrect. Given the information below, write a new domain and problem file in minimal PDDL that would solve the problem, starting from the failed step. Do not start from the initial state. Note the problem may not be solvable with the given information, in which case you should come up with a tangible subgoal for the problem OR infer what is missing.

Return a JSON object in the following format:
{{
  "domain file": ...,
  "problem file":...
}}

Domain Description:
{domain_description}

Problem Description:
{problem_description}

Common Sense Knowledge:
{statistics_knowledge}

Available Actions:
{available_actions}

Generated Domain File:
{previous_domain_file}

Generated Problem File:
{previous_problem_file}

Incorrect Found Plan:
{previous_plan}

Trajectory for Incorrect Plan:
{previous_trajectory}"""

grow_pddl_prompt_statistics = """You are a PDDL expert.

Using the following PDDL files and plan trajectory, modify the PDDL files based on the new observations. Note the problem may not be solvable with the given information, in which case you should come up with a tangible subgoal for the problem OR infer what is missing.

Return ONLY a JSON object in the following format:
{{
  "domain file": ...,
  "problem file":...
}}

Do NOT add extra text.

Domain Description:
{domain_description}

Problem Description:
{problem_description}

Common Sense Knowledge:
{statistics_knowledge}

Available Actions:
{available_actions}

Domain File:
{domain_file}

Problem File:
{problem_file}

Trajectory:
{trajectory}"""


\end{lstlisting}

\begin{lstlisting}[language=lisp, caption={Prompts for Hybrid Formalizer (fixed)}, label={lst:hybrid_formalizer_fixed}]
    initial_formalize_random = """You are a PDDL expert.

Given the information below, write the domain and problem files in minimal PDDL. Note that there might be information missing, in which case you should infer what is missing.

Return a JSON object in the following format:
{{
  "domain file": ...,
  "problem file":...
}}

Domain Description:
{domain_description}

Problem Description:
{problem_description}

Available Actions:
{available_actions}

Initial Observation:
{initial_observation}
"""    


initial_formalize_statistics = """You are a PDDL expert.

Given the information below, write the domain and problem files in minimal PDDL. Note that there might be information missing, in which case you should infer what is missing.

Return a JSON object in the following format:
{{
  "domain file": ...,
  "problem file":...
}}

Domain Description:
{domain_description}

Problem Description:
{problem_description}

Available Actions:
{available_actions}

Common Sense Knowledge:
{statistics_knowledge}

Initial Observation:
{initial_observation}
"""  


revision_random = """You are a PDDL expert.

The following example and generated PDDL contains an error. Using only the information below, revise the PDDL to remove the error. Note there might be information missing, in which case you should infer what is missing.

Return a JSON object in the following format:
{{
  "domain file": ...,
  "problem file":...
}}

Domain Description:
{domain_description}

Problem Description:
{problem_description}

Available Actions:
{available_actions}

Generated Domain File:
{previous_domain_file}

Generated Problem File:
{previous_problem_file}

Error Message:
{error_message}
"""     

revision_statistics = """You are a PDDL expert.

The following example and generated PDDL contains an error. Using only the information below, revise the PDDL to remove the error. Note there might be information missing, in which case you should infer what is missing.

Return a JSON object in the following format:
{{
  "domain file": ...,
  "problem file":...
}}

Domain Description:
{domain_description}

Problem Description:
{problem_description}

Available Actions:
{available_actions}

Common Sense Knowledge:
{statistics_knowledge}

Generated Domain File:
{previous_domain_file}

Generated Problem File:
{previous_problem_file}

Error Message:
{error_message}"""


re_formalize_random = """You are a PDDL expert.

In the following problem, the found plan from the domain and problem files is incorrect either due to incorrect assumptions or other mistakes. Given only the information below, write a new domain and problem file in minimal PDDL that would solve the problem, starting from the failed step. Do not start from the initial state. Note there might be information missing, in which case you should infer what is missing.

Return a JSON object in the following format:
{{
  "domain file": ...,
  "problem file":...
}}

Domain Description:
{domain_description}

Problem Description:
{problem_description}

Available Actions:
{available_actions}

Generated Domain File:
{previous_domain_file}

Generated Problem File:
{previous_problem_file}

Incorrect Found Plan:
{previous_plan}

Trajectory for Incorrect Plan:
{previous_trajectory}
"""


re_formalize_statistics = """You are a PDDL expert.

In the following problem, the found plan from the domain and problem files is incorrect either due to incorrect assumptions or other mistakes. Given only the information below, write a new domain and problem file in minimal PDDL that would solve the problem, starting from the failed step. Do not start from the initial state. Note there might be information missing, in which case you should infer what is missing.

Return a JSON object in the following format:
{{
  "domain file": ...,
  "problem file":...
}}

Domain Description:
{domain_description}

Problem Description:
{problem_description}

Available Actions:
{available_actions}

Common Sense Knowledge:
{statistics_knowledge}

Generated Domain File:
{previous_domain_file}

Generated Problem File:
{previous_problem_file}

Incorrect Found Plan:
{previous_plan}

Trajectory for Incorrect Plan:
{previous_trajectory}
"""


initial_observation_prompt = """You are a PDDL expert.

Using ONLY the information below, write the domain and problem file in minimal PDDL. Do NOT assume any information. Note the problem may not be solvable with the given information, in which case you should come up with a tangible subgoal for the problem.

Return ONLY a JSON object in the following format:
{{
  "domain file": ...,
  "problem file":...
}}

Do NOT add extra text.

Domain Description:
{domain_description}

Problem Description:
{problem_description}

Trajectory:
{trajectory}

Available Actions:
{available_actions}
"""

revision_prompt = """You are a PDDL expert.

The following generated PDDL contains an error. Using ONLY the information below, revise the PDDL to remove the error. Do NOT assume any information. Note the problem may not be solvable with the given information, in which case you should come up with a tangible subgoal for the problem.

Return ONLY a JSON object in the following format:
{{
  "domain file": ...,
  "problem file":...
}}

Do NOT add extra text.

Domain Description:
{domain_description}

Problem Description:
{problem_description}

Trajectory:
{trajectory}

Available Actions:
{available_actions}

Domain File:
{domain_file}

Problem File:
{problem_file}

Error:
{error}
"""            


re_generate_prompt = """You are a PDDL expert.

In the following problem, the found plan from the domain and problem files is incorrect. Given only the information below, write a new domain and problem file in minimal PDDL that would solve the problem, starting from the failed step. Do not start from the initial state. Do NOT assume any information. Note the problem may not be solvable with the given information, in which case you should come up with a tangible subgoal for the problem.

Return a JSON object in the following format:
{{
  "domain file": ...,
  "problem file":...
}}

Domain Description:
{domain_description}

Problem Description:
{problem_description}

Available Actions:
{available_actions}

Generated Domain File:
{previous_domain_file}

Generated Problem File:
{previous_problem_file}

Incorrect Found Plan:
{previous_plan}

Trajectory for Incorrect Plan:
{previous_trajectory}
"""          


grow_pddl_prompt = """You are a PDDL expert.

Using ONLY the following PDDL files and plan trajectory, modify the PDDL files based on the new observations. Do NOT assume any information. Note the problem may not be solvable with the given information, in which case you should come up with a tangible subgoal for the problem.

Return ONLY a JSON object in the following format:
{{
  "domain file": ...,
  "problem file":...
}}

Do NOT add extra text.

Domain Description:
{domain_description}

Problem Description:
{problem_description}

Available Actions:
{available_actions}

Domain File:
{domain_file}

Problem File:
{problem_file}

Trajectory:
{trajectory}"""            
\end{lstlisting}

\clearpage

\begin{table*}[t]
\centering
\small
\begin{tabular*}{\textwidth}{@{\extracolsep{\fill}}lrrrrrrrrrr@{}}
\toprule
 & \multicolumn{5}{c}{Without statistics} & \multicolumn{5}{c}{With statistics} \\
\cmidrule(lr){2-6}\cmidrule(l){7-11}
Method & Acc.$\uparrow$ & Att. & Fail.$\downarrow$ & API$\downarrow$ & Tok.$\downarrow$ & Acc.$\uparrow$ & Att. & Fail.$\downarrow$ & API$\downarrow$ & Tok.$\downarrow$ \\
\midrule
\multicolumn{11}{@{}l}{\textbf{Full observability}} \\
\multicolumn{11}{@{}l}{\textit{Planner}} \\
\quad Aggressive & 100 & 1.00 & 0.00 & 1.00 & 35 & 100 & 1.00 & 0.00 & 1.00 & 35 \\
\quad Conservative & 98 & 1.00 & 0.02 & 1.00 & 35 & -- & -- & -- & -- & -- \\
\quad Hybrid (free) & 100 & 1.00 & 0.00 & 1.00 & 35 & 94 & 1.00 & 0.06 & 1.00 & 35 \\
\quad Hybrid (fixed) & 100 & 1.00 & 0.00 & 1.00 & 35 & 100 & 1.00 & 0.00 & 1.00 & 34 \\
\multicolumn{11}{@{}l}{\textit{Formalizer}} \\
\quad Aggressive & 100 & 1.00 & 0.00 & 1.02 & 485 & 94 & 1.00 & 0.06 & 1.01 & 468 \\
\quad Conservative & 69 & 1.00 & 0.31 & 1.77 & 727 & -- & -- & -- & -- & -- \\
\quad Hybrid (free) & 93 & 1.00 & 0.07 & 1.01 & 466 & 90 & 1.00 & 0.10 & 1.01 & 456 \\
\quad Hybrid (fixed) & 100 & 1.00 & 0.00 & 1.01 & 483 & 92 & 1.00 & 0.08 & 1.04 & 490 \\
\midrule
\multicolumn{11}{@{}l}{\textbf{Missing all connections}} \\
\multicolumn{11}{@{}l}{\textit{Planner}} \\
\quad Aggressive & 99 & 4.09 & 3.10 & 4.09 & 111 & 100 & 2.47 & 1.47 & 2.47 & 86 \\
\quad Conservative & 100 & 1.03 & 0.03 & 4.23 & 52 & -- & -- & -- & -- & -- \\
\quad Hybrid (free) & 100 & 2.08 & 1.08 & 4.07 & 73 & 100 & 2.32 & 1.32 & 2.48 & 80 \\
\quad Hybrid (fixed) & 100 & 2.71 & 1.71 & 4.22 & 90 & 100 & 2.01 & 1.01 & 2.71 & 71 \\
\multicolumn{11}{@{}l}{\textit{Formalizer}} \\
\quad Aggressive & 99 & 4.10 & 3.11 & 4.20 & 1{,}515 & 97 & 3.00 & 2.03 & 3.10 & 1{,}211 \\
\quad Conservative & 99 & 2.58 & 1.59 & 5.84 & 1{,}947 & -- & -- & -- & -- & -- \\
\quad Hybrid (free) & 97 & 4.27 & 3.30 & 4.60 & 1{,}500 & 99 & 2.93 & 1.94 & 3.11 & 1{,}108 \\
\quad Hybrid (fixed) & 94 & 4.62 & 3.68 & 6.31 & 2{,}212 & 100 & 3.24 & 2.24 & 4.23 & 1{,}632 \\
\bottomrule
\end{tabular*}
\caption{Coin Collector results for DeepSeek-Flash (run 1). Acc.\ is planning accuracy (\%), Att.\ the average number of attempts per problem, Fail.\ the average number of failed attempts per problem (attempts minus accuracy), API the average number of API calls per problem, and Tok.\ the average number of output tokens per problem. Conservative methods are not given statistics (--). All rows use 100 problems.}
\label{tab:cc-deepseek-run1}
\end{table*}

\begin{table*}[t]
\centering
\small
\begin{tabular*}{\textwidth}{@{\extracolsep{\fill}}lrrrrrrrrrr@{}}
\toprule
 & \multicolumn{5}{c}{Without statistics} & \multicolumn{5}{c}{With statistics} \\
\cmidrule(lr){2-6}\cmidrule(l){7-11}
Method & Acc.$\uparrow$ & Att. & Fail.$\downarrow$ & API$\downarrow$ & Tok.$\downarrow$ & Acc.$\uparrow$ & Att. & Fail.$\downarrow$ & API$\downarrow$ & Tok.$\downarrow$ \\
\midrule
\multicolumn{11}{@{}l}{\textbf{Full observability}} \\
\multicolumn{11}{@{}l}{\textit{Planner}} \\
\quad Aggressive & 100 & 1.00 & 0.00 & 1.00 & 35 & 98 & 1.00 & 0.02 & 1.00 & 34 \\
\quad Conservative & 100 & 1.00 & 0.00 & 1.00 & 35 & -- & -- & -- & -- & -- \\
\quad Hybrid (free) & 100 & 1.00 & 0.00 & 1.00 & 35 & 99 & 1.00 & 0.01 & 1.00 & 34 \\
\quad Hybrid (fixed) & 100 & 1.00 & 0.00 & 1.00 & 35 & 99 & 1.00 & 0.01 & 1.00 & 34 \\
\multicolumn{11}{@{}l}{\textit{Formalizer}} \\
\quad Aggressive & 98 & 1.00 & 0.02 & 1.00 & 438 & 91 & 1.00 & 0.09 & 1.00 & 442 \\
\quad Conservative & 95 & 1.00 & 0.05 & 1.67 & 634 & -- & -- & -- & -- & -- \\
\quad Hybrid (free) & 73 & 1.00 & 0.27 & 1.00 & 420 & 76 & 1.00 & 0.24 & 1.01 & 427 \\
\quad Hybrid (fixed) & 96 & 1.00 & 0.04 & 1.01 & 440 & 88 & 1.00 & 0.12 & 1.02 & 454 \\
\midrule
\multicolumn{11}{@{}l}{\textbf{Missing all connections}} \\
\multicolumn{11}{@{}l}{\textit{Planner}} \\
\quad Aggressive & 99 & 3.68 & 2.69 & 3.68 & 106 & 100 & 2.51 & 1.51 & 2.51 & 85 \\
\quad Conservative & 100 & 1.09 & 0.09 & 4.43 & 54 & -- & -- & -- & -- & -- \\
\quad Hybrid (free) & 100 & 3.16 & 2.16 & 3.63 & 100 & 100 & 2.47 & 1.47 & 2.47 & 84 \\
\quad Hybrid (fixed) & 100 & 2.65 & 1.65 & 3.92 & 80 & 100 & 2.09 & 1.09 & 2.90 & 74 \\
\multicolumn{11}{@{}l}{\textit{Formalizer}} \\
\quad Aggressive & 100 & 3.89 & 2.89 & 3.90 & 1{,}432 & 99 & 3.10 & 2.11 & 3.11 & 1{,}248 \\
\quad Conservative & 100 & 1.29 & 0.29 & 4.91 & 1{,}572 & -- & -- & -- & -- & -- \\
\quad Hybrid (free) & 100 & 3.70 & 2.70 & 3.70 & 1{,}248 & 99 & 3.25 & 2.26 & 3.26 & 1{,}241 \\
\quad Hybrid (fixed) & 100 & 3.34 & 2.34 & 4.98 & 1{,}711 & 100 & 2.82 & 1.82 & 4.14 & 1{,}557 \\
\bottomrule
\end{tabular*}
\caption{Coin Collector results for Gemini-3-Flash (run 1). Acc.\ is planning accuracy (\%), Att.\ the average number of attempts per problem, Fail.\ the average number of failed attempts per problem (attempts minus accuracy), API the average number of API calls per problem, and Tok.\ the average number of output tokens per problem. Conservative methods are not given statistics (--). All rows use 100 problems.}
\label{tab:cc-gemini-run1}
\end{table*}

\begin{table*}[t]
\centering
\small
\begin{tabular*}{\textwidth}{@{\extracolsep{\fill}}lrrrrrrrrrr@{}}
\toprule
 & \multicolumn{5}{c}{Without statistics} & \multicolumn{5}{c}{With statistics} \\
\cmidrule(lr){2-6}\cmidrule(l){7-11}
Method & Acc.$\uparrow$ & Att. & Fail.$\downarrow$ & API$\downarrow$ & Tok.$\downarrow$ & Acc.$\uparrow$ & Att. & Fail.$\downarrow$ & API$\downarrow$ & Tok.$\downarrow$ \\
\midrule
\multicolumn{11}{@{}l}{\textbf{Full observability}} \\
\multicolumn{11}{@{}l}{\textit{Planner}} \\
\quad Aggressive & 100 & 1.00 & 0.00 & 1.00 & 35 & 98 & 1.00 & 0.02 & 1.00 & 34 \\
\quad Conservative & 98 & 1.00 & 0.02 & 1.00 & 35 & -- & -- & -- & -- & -- \\
\quad Hybrid (free) & 97 & 1.00 & 0.03 & 1.00 & 35 & 98 & 1.00 & 0.02 & 1.00 & 35 \\
\quad Hybrid (fixed) & 99 & 1.00 & 0.01 & 1.00 & 35 & 99 & 1.00 & 0.01 & 1.00 & 34 \\
\multicolumn{11}{@{}l}{\textit{Formalizer}} \\
\quad Aggressive & 83 & 1.00 & 0.17 & 1.67 & 831 & 83 & 1.00 & 0.17 & 1.69 & 816 \\
\quad Conservative & 80 & 1.00 & 0.20 & 1.93 & 821 & -- & -- & -- & -- & -- \\
\quad Hybrid (free) & 78 & 1.00 & 0.22 & 1.74 & 837 & 72 & 1.00 & 0.28 & 1.88 & 885 \\
\quad Hybrid (fixed) & 84 & 1.00 & 0.16 & 1.69 & 843 & 77 & 1.00 & 0.23 & 1.80 & 872 \\
\midrule
\multicolumn{11}{@{}l}{\textbf{Missing all connections}} \\
\multicolumn{11}{@{}l}{\textit{Planner}} \\
\quad Aggressive & 100 & 3.61 & 2.61 & 3.61 & 93 & 100 & 2.62 & 1.62 & 2.62 & 89 \\
\quad Conservative & 100 & 1.00 & 0.00 & 4.45 & 53 & -- & -- & -- & -- & -- \\
\quad Hybrid (free) & 100 & 2.18 & 1.18 & 3.96 & 69 & 100 & 2.39 & 1.39 & 2.50 & 82 \\
\quad Hybrid (fixed) & 100 & 2.63 & 1.63 & 4.11 & 79 & 100 & 2.06 & 1.06 & 2.99 & 74 \\
\multicolumn{11}{@{}l}{\textit{Formalizer}} \\
\quad Aggressive & 75 & 3.63 & 2.88 & 4.46 & 1{,}464 & 73 & 2.83 & 2.10 & 3.59 & 1{,}277 \\
\quad Conservative & 76 & 1.08 & 0.32 & 4.68 & 1{,}477 & -- & -- & -- & -- & -- \\
\quad Hybrid (free) & 76 & 3.61 & 2.85 & 4.70 & 1{,}369 & 70 & 2.63 & 1.93 & 3.83 & 1{,}188 \\
\quad Hybrid (fixed) & 73 & 3.32 & 2.59 & 5.58 & 1{,}861 & 71 & 2.07 & 1.36 & 3.97 & 1{,}531 \\
\bottomrule
\end{tabular*}
\caption{Coin Collector results for Qwen3.8-27B (run 1). Acc.\ is planning accuracy (\%), Att.\ the average number of attempts per problem, Fail.\ the average number of failed attempts per problem (attempts minus accuracy), API the average number of API calls per problem, and Tok.\ the average number of output tokens per problem. Conservative methods are not given statistics (--). All rows use 100 problems.}
\label{tab:cc-qwen3.8-run1}
\end{table*}

\begin{table*}[t]
\centering
\small
\begin{tabular*}{\textwidth}{@{\extracolsep{\fill}}lrrrrrrrrrr@{}}
\toprule
 & \multicolumn{5}{c}{Without statistics} & \multicolumn{5}{c}{With statistics} \\
\cmidrule(lr){2-6}\cmidrule(l){7-11}
Method & Acc.$\uparrow$ & Att. & Fail.$\downarrow$ & API$\downarrow$ & Tok.$\downarrow$ & Acc.$\uparrow$ & Att. & Fail.$\downarrow$ & API$\downarrow$ & Tok.$\downarrow$ \\
\midrule
\multicolumn{11}{@{}l}{\textbf{Full observability}} \\
\multicolumn{11}{@{}l}{\textit{Planner}} \\
\quad Aggressive & 100 & 1.00 & 0.00 & 1.00 & 35 & 100 & 1.00 & 0.00 & 1.00 & 35 \\
\quad Conservative & 100 & 1.00 & 0.00 & 1.00 & 35 & -- & -- & -- & -- & -- \\
\quad Hybrid (free) & 100 & 1.00 & 0.00 & 1.00 & 35 & 99 & 1.00 & 0.01 & 1.00 & 35 \\
\quad Hybrid (fixed) & 100 & 1.00 & 0.00 & 1.00 & 35 & 100 & 1.00 & 0.00 & 1.00 & 35 \\
\multicolumn{11}{@{}l}{\textit{Formalizer}} \\
\quad Aggressive & 98 & 1.00 & 0.02 & 1.03 & 494 & 96 & 1.00 & 0.04 & 1.00 & 472 \\
\quad Conservative & 75 & 1.00 & 0.25 & 1.73 & 717 & -- & -- & -- & -- & -- \\
\quad Hybrid (free) & 95 & 1.00 & 0.05 & 1.03 & 479 & 93 & 1.00 & 0.07 & 1.03 & 473 \\
\quad Hybrid (fixed) & 99 & 1.00 & 0.01 & 1.00 & 478 & 91 & 1.00 & 0.09 & 1.00 & 469 \\
\midrule
\multicolumn{11}{@{}l}{\textbf{Missing all connections}} \\
\multicolumn{11}{@{}l}{\textit{Planner}} \\
\quad Aggressive & 100 & 3.98 & 2.98 & 3.98 & 107 & 100 & 2.32 & 1.32 & 2.32 & 78 \\
\quad Conservative & 100 & 1.01 & 0.01 & 4.28 & 55 & -- & -- & -- & -- & -- \\
\quad Hybrid (free) & 100 & 2.15 & 1.15 & 4.00 & 72 & 100 & 2.31 & 1.31 & 2.45 & 78 \\
\quad Hybrid (fixed) & 100 & 2.59 & 1.59 & 3.74 & 78 & 100 & 2.06 & 1.06 & 2.81 & 76 \\
\multicolumn{11}{@{}l}{\textit{Formalizer}} \\
\quad Aggressive & 98 & 4.27 & 3.29 & 4.35 & 1{,}594 & 100 & 2.93 & 1.93 & 3.03 & 1{,}166 \\
\quad Conservative & 96 & 2.87 & 1.91 & 6.39 & 2{,}127 & -- & -- & -- & -- & -- \\
\quad Hybrid (free) & 98 & 4.25 & 3.27 & 4.42 & 1{,}443 & 97 & 2.94 & 1.97 & 3.17 & 1{,}133 \\
\quad Hybrid (fixed) & 93 & 4.71 & 3.78 & 6.47 & 2{,}283 & 96 & 3.51 & 2.55 & 4.57 & 1{,}739 \\
\bottomrule
\end{tabular*}
\caption{Coin Collector results for DeepSeek-Flash (run 2). Acc.\ is planning accuracy (\%), Att.\ the average number of attempts per problem, Fail.\ the average number of failed attempts per problem (attempts minus accuracy), API the average number of API calls per problem, and Tok.\ the average number of output tokens per problem. Conservative methods are not given statistics (--). All rows use 100 problems.}
\label{tab:cc-deepseek-run2}
\end{table*}

\begin{table*}[t]
\centering
\small
\begin{tabular*}{\textwidth}{@{\extracolsep{\fill}}lrrrrrrrrrr@{}}
\toprule
 & \multicolumn{5}{c}{Without statistics} & \multicolumn{5}{c}{With statistics} \\
\cmidrule(lr){2-6}\cmidrule(l){7-11}
Method & Acc.$\uparrow$ & Att. & Fail.$\downarrow$ & API$\downarrow$ & Tok.$\downarrow$ & Acc.$\uparrow$ & Att. & Fail.$\downarrow$ & API$\downarrow$ & Tok.$\downarrow$ \\
\midrule
\multicolumn{11}{@{}l}{\textbf{Full observability}} \\
\multicolumn{11}{@{}l}{\textit{Planner}} \\
\quad Aggressive & 100 & 1.00 & 0.00 & 1.00 & 35 & 100 & 1.00 & 0.00 & 1.00 & 35 \\
\quad Conservative & 98 & 1.00 & 0.02 & 1.00 & 35 & -- & -- & -- & -- & -- \\
\quad Hybrid (free) & 100 & 1.00 & 0.00 & 1.00 & 35 & 96 & 1.00 & 0.04 & 1.00 & 35 \\
\quad Hybrid (fixed) & 100 & 1.00 & 0.00 & 1.00 & 35 & 100 & 1.00 & 0.00 & 1.00 & 35 \\
\multicolumn{11}{@{}l}{\textit{Formalizer}} \\
\quad Aggressive & 99 & 1.00 & 0.01 & 1.01 & 486 & 94 & 1.00 & 0.06 & 1.00 & 476 \\
\quad Conservative & 69 & 1.00 & 0.31 & 1.63 & 679 & -- & -- & -- & -- & -- \\
\quad Hybrid (free) & 93 & 1.00 & 0.07 & 1.01 & 470 & 90 & 1.00 & 0.10 & 1.01 & 456 \\
\quad Hybrid (fixed) & 100 & 1.00 & 0.00 & 1.02 & 490 & 86 & 1.00 & 0.14 & 1.00 & 470 \\
\midrule
\multicolumn{11}{@{}l}{\textbf{Missing all connections}} \\
\multicolumn{11}{@{}l}{\textit{Planner}} \\
\quad Aggressive & 99 & 3.81 & 2.82 & 3.81 & 100 & 100 & 2.40 & 1.40 & 2.40 & 81 \\
\quad Conservative & 100 & 1.00 & 0.00 & 4.33 & 53 & -- & -- & -- & -- & -- \\
\quad Hybrid (free) & 100 & 2.21 & 1.21 & 4.10 & 74 & 100 & 2.28 & 1.28 & 2.47 & 78 \\
\quad Hybrid (fixed) & 100 & 2.61 & 1.61 & 4.03 & 83 & 100 & 2.07 & 1.07 & 2.82 & 75 \\
\multicolumn{11}{@{}l}{\textit{Formalizer}} \\
\quad Aggressive & 97 & 4.52 & 3.55 & 4.65 & 1{,}682 & 98 & 2.98 & 2.00 & 3.00 & 1{,}168 \\
\quad Conservative & 100 & 2.67 & 1.67 & 6.07 & 2{,}012 & -- & -- & -- & -- & -- \\
\quad Hybrid (free) & 99 & 4.02 & 3.03 & 4.30 & 1{,}352 & 99 & 2.91 & 1.92 & 3.04 & 1{,}085 \\
\quad Hybrid (fixed) & 94 & 4.56 & 3.62 & 6.05 & 2{,}109 & 96 & 3.46 & 2.50 & 4.42 & 1{,}712 \\
\bottomrule
\end{tabular*}
\caption{Coin Collector results for DeepSeek-Flash (run 3). Acc.\ is planning accuracy (\%), Att.\ the average number of attempts per problem, Fail.\ the average number of failed attempts per problem (attempts minus accuracy), API the average number of API calls per problem, and Tok.\ the average number of output tokens per problem. Conservative methods are not given statistics (--). All rows use 100 problems.}
\label{tab:cc-deepseek-run3}
\end{table*}

\begin{table*}[t]
\centering
\small
\begin{tabular*}{\textwidth}{@{\extracolsep{\fill}}lrrrrr@{}}
\toprule
Method & Acc.$\uparrow$ & Att. & Fail.$\downarrow$ & API$\downarrow$ & Tok.$\downarrow$ \\
\midrule
\multicolumn{6}{@{}l}{\textbf{DeepSeek-Flash}} \\
\quad Aggressive planner & 98 & 3.81 & 2.83 & 3.81 & 146 \\
\quad Conservative planner & 100 & 2.33 & 1.33 & 5.69 & 159 \\
\quad Aggressive formalizer & 75 & 11.40 & 10.65 & 12.34 & 10{,}240 \\
\quad Conservative formalizer & 63 & 12.21 & 11.58 & 24.41 & 16{,}718 \\
\midrule
\multicolumn{6}{@{}l}{\textbf{Qwen3.8-27B}} \\
\quad Aggressive planner & 100 & 3.67 & 2.67 & 3.67 & 137 \\
\quad Conservative planner & 100 & 1.14 & 0.14 & 5.89 & 178 \\
\quad Aggressive formalizer & 26 & 3.33 & 3.07 & 6.20 & 6{,}609 \\
\quad Conservative formalizer & 30 & 5.59 & 5.29 & 18.26 & 11{,}371 \\
\bottomrule
\end{tabular*}
\caption{ALFWorld results for DeepSeek-Flash and Qwen3.8-27B, 100 problems per method, without statistics. Acc.\ is planning accuracy (\%), Att.\ the average number of attempts per problem, Fail.\ the average number of failed attempts per problem (attempts minus accuracy), API the average number of API calls per problem, and Tok.\ the average number of output tokens per problem. All rows use 100 problems.}
\label{tab:alfworld-all}
\end{table*}

\end{document}